%% file: arxiv_upload_11_2026.tex
\documentclass[lettersize,journal]{IEEEtran}
\usepackage{amsmath,amsfonts}
\usepackage{algorithmic}
\usepackage{algorithm}
\usepackage{array}
\usepackage[font=normalsize]{subcaption}
\usepackage{textcomp}
\usepackage{stfloats}
\usepackage{url}
\usepackage{verbatim}
\usepackage{graphicx}
\usepackage{cite}
\usepackage{multirow}
\usepackage{xr}
\usepackage{hyperref}
\begin{document}

\title{Rethinking ${GA^2Ms}$: performance and out-of-distribution generalisation}

\author{George Dimitriadis*, Damian Machlanski, Paul Clarke, Spyridon Samothrakis*
\thanks{George Dimitriadis is with the Computer Science and Electronic Engineering Department of the University of Essex. email: gd25222@essex.ac.uk}
\thanks{Damian Machlanski is with School of Engineering at the University of Edinburgh and with Causality in Healthcare AI Hub (CHAI). email: d.machlanski@ed.ac.uk}
\thanks{Paul Clarke is with the Institute for Social and Economic Research and the ESRC Research Centre on Micro-Social Change (MiSoC), Univesrity of Essex. email:pclarke@essex.ac.uk}
\thanks{Spyridon Samothrakis is with the Computer Science and Electronic Engineering Department of the University of Essex. email: ssamot@essex.ac.uk}
\thanks{The code used to generate this article can be found here \href{https://github.com/georgedimitriadis/rethinking_GA2Ms}{https://github.com/georgedimitriadis/rethinking\_GA2Ms}.}%
\thanks{GD did part of the code development, the analysis and the paper authoring. DM contributed in the conceptualisation of the $CoGA^2M$ model and did the original TALENT experiments. PC added the statistical approach to the $CoGA^2M$ model. SS did most of the $CoGA^2M$ code development, helped in its conceptualisation and oversaw the project.}}



\maketitle

\begin{abstract}
Generalised Additive Models (GAMs) are a cornerstone of interpretable machine learning, offering transparent predictions through structured additive decompositions. Despite advances from classical GAMs to modern neural and spline-based variants, two notable gaps remain. First, comprehensive benchmarking of spline-based GAMs on large-scale tabular learning tasks—the dominant modality in applied settings—is lacking. Second, their robustness under distribution shift, a setting where structured inductive biases should confer advantages, remains unexplored. We propose a multi-resolution, higher-order Composite Generalised Additive Model (${CoGA^2M}$) and evaluate it on the TALENT benchmark suite. CoGA²M ranks in the top third across all baselines (29 for regression, 32 for classification), establishing competitive performance with state-of-the-art tabular methods. We further extend TALENT with out-of-distribution (OOD) splits exhibiting train-test distributional shift and benchmark ${CoGA^2M}$ against LightGBM, a consistently top-performing model on TALENT. Under distribution shift, ${CoGA^2M}$ outperforms LightGBM, demonstrating a clear structural advantage. Our results position spline-based GAMs as strong contenders for tabular prediction and reveal a previously unrecognised benefit of additive structure in the OOD regime.
\end{abstract}


\section{Introduction}
Generalised Additive Models (GAMs), first proposed by Hastie and Tibshirani \cite{hastieGeneralizedAdditiveModels}, have long occupied a distinctive niche in machine learning: they offer interpretable, structured predictions by modelling the response as a sum of smooth functions of individual features. Over the past three decades, a rich family of extensions has emerged, ranging from penalised spline methods \cite{eilersFlexibleSmoothingBsplines1996, woodThinPlateRegression2003} to higher-order interaction models (${GA^2Ms}$) \cite{louAccurateIntelligibleModels2013}, constrained variants \cite{watanabeConstrainedGeneralizedAdditive2021}, boosting-based approaches such as Explainable Boosting Machines \cite{watanabeConstrainedGeneralizedAdditive2021}, and neural network–based formulations \cite{agarwalNeuralAdditiveModels2021, yangGAMINetExplainableNeural2021, yangEnhancingExplainabilityNeural2021, kruschelChallengingPerformanceInterpretabilityTradeOff2025}. A unifying theme across this line of work is the emphasis on explainability, and recent research has demonstrated that interpretability need not come at the cost of predictive performance \cite{kruschelChallengingPerformanceInterpretabilityTradeOff2025}. This finding is encouraging, yet the empirical evaluations in the GAM literature have largely been confined to standard in-distribution (ID) settings.

The study of out-of-distribution (OOD) generalisation has received considerable attention in other areas of machine learning, with theoretical foundations in domain adaptation and covariate shift \cite{ben-davidTheoryLearningDifferent2010}, and a range of algorithmic families proposed to improve robustness to distribution shift \cite{arjovskyInvariantRiskMinimization2019, sagawaDistributionallyRobustNeural2019, kruegerOutofDistributionGeneralizationRisk2021, ganinDomainAdversarialTrainingNeural2017, zhangMixupEmpiricalRisk2018}. On the evaluation side, substantial benchmarking efforts have been directed at image \cite{gulrajaniSearchLostDomain2020, kohWILDSBenchmarkIntheWild2021}, text \cite{yuanRevisitingOutofdistributionRobustness2023, yangGLUEXEvaluatingNatural2023} and general multi-modal settings \cite{yeOoDBenchQuantifyingUnderstanding2026}. However, the vast majority of this research — both methodological and empirical — has concentrated on image and text domains. The tabular data setting, which remains the most prevalent data modality in practice, has received comparatively little attention. The TableShift benchmark \cite{gardnerBenchmarkingDistributionShift2023} represents a notable recent effort, but even it focuses primarily on comparing deep learning architectures and generic robustness methods, rather than examining how the inductive biases of specific model families — such as the smoothness and additivity priors of GAMs — affect OOD behaviour. To our knowledge, there has been no systematic effort to examine the behaviour of GAMs and their variants under OOD conditions — a critical gap given that real-world deployment routinely exposes models to data that diverge from the training distribution. This is a critical obstacle to safe and reliable deployment in high-stakes environments \cite{subbaswamyDevelopmentDeploymentDataset2020, subasriDetectingRemediatingHarmful2025}.

This gap motivates the present work. The smooth, structured nature of spline-based models suggests they may extrapolate more gracefully than flexible black-box alternatives when faced with diversity shifts — regions of input space unseen during training where the underlying data-generating process remains unchanged. Yet this hypothesis has not been tested empirically. Relatedly, while ${GA^2M}$-family models have been shown to be competitive on standard benchmarks, no existing study places them on an equal footing with state-of-the-art methods across a large-scale tabular evaluation while simultaneously probing their OOD behaviour.

To address both gaps, we make two contributions. First, we introduce a multi-resolution, higher-order, P-Spline, Composite Generalised Additive Model (${CoGA^2M}$) that generates splines at multiple resolutions and arities, combining linear and cubic Hermite basis functions into a single differentiable layer whose parameters — including the regularisation strength — are learned end-to-end via backpropagation. We evaluate ${CoGA^2M}$ on the full TALENT benchmark \cite{liuTalentTabularAnalytics} comprising 300 tabular datasets, demonstrating that it ranks in the top third of all 30+ models across regression, binary classification, and multi-class classification tasks, with performance comparable to leading methods such as modernNCA, LightGBM, XGBoost, realmlp, tabr and CatBoost. Second, we develop an adversarial OOD dataset generation procedure that carves compact, connected regions out of a dataset's interior to create diversity-shifted train–test splits. Using an evolutionary strategy (CMA-ES) \cite{hansenCMAEvolutionStrategy2023} in an adversarial loop with learner models, we produce progressively harder OOD splits for 40 regression datasets. We then compare ${CoGA^2M}$ against LightGBM on these OOD splits, finding that ${CoGA^2M}$ exhibits a statistically significant smaller drop in performance when moving from ID to OOD data (bootstrap p = 0.006), suggesting that the smooth, spline-based inductive bias of $GA^2M$-family models confers a meaningful advantage in diversity-shift OOD scenarios. In this study we focus exclusively on diversity shift.

The remainder of the paper is organised as follows. Section II reviews related work on GAM-family models and OOD generalisation. Section III describes our methods, including the ${CoGA^2M}$ architecture, the adversarial OOD dataset generation procedure, and the experimental setup. Section IV presents results on both the TALENT benchmark and the OOD comparison. Section V concludes.

\section{Background}
\label{related_work}
\subsection{Generalised Additive Models and their extensions}
Generalised Additive Models were introduced by Hastie and Tibshirani \cite{hastieGeneralizedAdditiveModels} with a local scoring algorithm to automatically determine the appropriate number of knots. Penalised B-splines (P-Splines) \cite{eilersFlexibleSmoothingBsplines1996} introduced efficient roughness penalties for controlling smoothness, while penalised thin plate regression splines with shrinkage (TPSplines) \cite{woodThinPlateRegression2003} offered a more flexible alternative that avoids manual knot placement. The move to higher-order interactions gave rise to ${GA^2M}$s \cite{louAccurateIntelligibleModels2013}, which augment the additive structure with pairwise feature terms, substantially increasing expressiveness while preserving a degree of interpretability. ${GA^2M}$s introduced the FAST measure to select the most appropriate feature combinations. Further developments include ${CGA^2M+}$s \cite{watanabeConstrainedGeneralizedAdditive2021}, which enforce monotonicity constraints on selected features, and Explainable Boosting Machines (EBMs) \cite{watanabeConstrainedGeneralizedAdditive2021}, which use a cyclic gradient boosting procedure to fit GAM-style models at scale.
More recently, neural network–based variants have gained traction. Neural Additive Models (NAMs) \cite{agarwalNeuralAdditiveModels2021} replace traditional splines with learned neural sub-networks, GAMI-Net \cite{yangGAMINetExplainableNeural2021} introduces structured interaction selection into neural GAMs, Explainable Neural Networks (ExNNs) \cite{yangEnhancingExplainabilityNeural2021} impose architectural constraints for transparency and interpretability, and Interpretable Generalised Additive Neural Networks (IGANNs) \cite{kruschelChallengingPerformanceInterpretabilityTradeOff2025} combine neural flexibility with additive structure. Kruschel et al. \cite{kruschelChallengingPerformanceInterpretabilityTradeOff2025} present an initial evaluation of GAM based models showing that interpretable machine learning models can challenge the conventional performance–interpretability trade-off. A 20 dataset (10 binary classification, 10 regression) benchmark was constructed and used to compare 7 GAM based models, two classical, interpretable models (linear/logistic regression and decision-trees) and 5 black-box models (random forests, XGBoost \cite{chenXGBoostScalableTree2016}, CatBoost \cite{prokhorenkovaCatBoostUnbiasedBoosting2018}, Multi Layer Perceptron (MLP) and TabNet \cite{arikTabNetAttentiveInterpretable2021}). The GAM based approached achieved competitive accuracy relative to the black-box alternatives while showing better interpretability, as defined by 6 measures (additivity, sparsity, linearity, smoothness, monotonicity and visualizability).  

\subsection{Out-of-distribution generalisation}
The theoretical foundations of OOD generalisation were laid by work on domain adaptation and covariate shift \cite{ben-davidTheoryLearningDifferent2010}, which formalised how discrepancies between training and test distributions affect generalisation bounds. Building on these ideas, several algorithmic families have been proposed to improve robustness to distribution shift. Invariant Risk Minimisation (IRM) \cite{arjovskyInvariantRiskMinimization2019} seeks to learn feature representations that support an optimal predictor across data drawn from any possible distribution (usually known as training domains, here referred to as environments) which, importantly, supports an underlying, invariant across environments, causal mechanism. Those include environments not seen during learning. Given training data from an adequate number of environments the method encourages invariance to spurious correlations and optimality in OOD scenarios. Distributionally Robust Optimisation (DRO) \cite{sagawaDistributionallyRobustNeural2019} takes a complementary approach, minimising worst-case risk over a set of predefined groups or perturbation sets. Again, this method targets datasets where spurious correlations exist in small sub-sets which although averaged over in the training regime can become prominent in the test data. The method relies on heavy regularisation techniques to allow the model to focus its risk minimization on the worse cases of such sub-groups. Risk Extrapolation (REx) \cite{kruegerOutofDistributionGeneralizationRisk2021} extends DRO by extrapolating beyond the convex hull of training risks to achieve robustness against more severe shifts. The concept here is that if a method achieves both an almost minimum risk (kept equal across several domains) then it will be able to extrapolate to unseen and more severe distribution shifts far away from the ones represented by the training domains. Domain-adversarial methods \cite{ganinDomainAdversarialTrainingNeural2017} learn representations that are indistinguishable across domains. Data augmentation strategies such as Mixup \cite{zhangMixupEmpiricalRisk2018} and its domain-generalisation variants create virtual data points that are linear interpolations between real points are designed to improve generalisation, model calibration and robustness against adversarial examples. Yet, and despite these methods not targeting OOD data or dealing with multiple domains they also deal better with OOD datasets, especially distribution shifts where the test data are drawn from a separate part of the distribution compared to the train set, simply because by accessing virtual data the model is forced to learn something about regions where real data do not exist. A more extensive account of the current state of the OOD literature can be found in a recent survey \cite{zhouDomainGeneralizationSurvey2023}.

\subsection{OOD benchmarks and evaluation}
On the evaluation side, substantial efforts have been directed at creating standardised benchmarks to assess OOD methods. DomainBed \cite{gulrajaniSearchLostDomain2020} established a unified test-bed for domain generalisation in computer vision, encompassing seven multi-domain image datasets and multiple algorithms — and notably demonstrated that carefully implemented Empirical Risk Minimisation (ERM) can match or exceed many specialised OOD methods. WILDS \cite{kohWILDSBenchmarkIntheWild2021} broadened the scope to real-world distribution shifts across diverse data modalities including medical imaging, satellite imagery, and natural language. The OoDBench+ framework \cite{yeOoDBenchQuantifyingUnderstanding2026} proposed a particularly useful taxonomy, distinguishing between two principal types of domain shift: diversity shift, where the test data is drawn from a region of the input space unseen during training while the underlying data-generating process remains unchanged, and correlation shift, where spurious correlations present in the training data are altered or reversed at test time (see Fig. \ref{fig:id_ood_schematic} for a visual representation of these ideas). Under a causal framework diversity shift is a generalisation of covariate shift (where the distribution of X changes but not the distribution of the labels given X). Correlation shift on the other hand can be considered as a latent shift \cite{tsaiProxyMethodsDomain2024} since the spurious correlations can be attributed to a correlation of some driving latent variable with both some features and the labels. In the natural language processing domain, benchmarks such as BOSS \cite{yuanRevisitingOutofdistributionRobustness2023} and GLUE-X \cite{yangGLUEXEvaluatingNatural2023}, have further advanced the field. BOSS leverages Simple Contrastive Sentence Embeddings (SimCSE) to calculate semantic similarity between phrases and is in this way able to evaluate OOD distances between language datasets. It then utilises this metric to construct maximally challenging train–test splits. GLUE-X evaluates language understanding models on a series of language benchmarks chosen for their OOD focus. 

\input{Fig1}

\subsection{Tabular data under distribution shift}
Despite this progress, the vast majority of OOD research — both methodological and empirical — has concentrated on image and text domains. The tabular data setting, which remains the most prevalent data modality in practice, has received comparatively little attention. The TableShift benchmark \cite{gardnerBenchmarkingDistributionShift2023} represents a notable recent effort, providing 15 curated binary classification tasks under real-world distribution shifts and revealing that the robustness of tabular models to distribution shift is poorly understood. Yet even TableShift focuses primarily on comparing deep learning architectures and generic robustness methods, rather than examining how the inductive biases of specific model families — such as the smoothness and additivity priors of GAMs — affect OOD behaviour. More broadly, relatively few efforts have focused on rigorously evaluating established tabular algorithms under controlled distribution shifts, and the question of whether interpretable models generalise differently from black-box ones in OOD settings remains largely unanswered.

\section{Methods}
\subsection{Multi-resolution composite P-Splines}
\label{methods_splines}
We develop an extension to the P-Splines algorithm \cite{eilersFlexibleSmoothingBsplines1996} in which for each feature, we generate splines with different number of knots (spline resolution) and with different arity (number of features as input). For each resolution and arity we create both a linear and a cubic spline. The results of all the splines acting on their appropriate feature values are then concatenated into the final result of the Splines' layer. This output is then passed into a Linear layer which combines the Spline's layer output into either a single regression result or a set of probabilities for classification problems. Both the parameters \(\theta\) that parametrise each spline and the weights \(w\) for the final linear composition are learned.\\
For arity bigger than one, the generated splines are the product of the equivalent univariate splines for each feature. So for arity two we generate a linear and a cubic spline for each one of the feature pairs as the product of the linear and cubic splines of each individual feature. For arity three, that product would arise from the three univariate splines, and so on. In this work we use up to arity two. We have found that arity one does not do as well in all of the datasets we tried. Arity three and above models are both prohibitively expensive and from initial experiments did not seem to provide any significant improvements.\\
\input{Fig2}
\subsubsection{Spline basis functions}
For input \(x\in[-1,1]\) mapped to a scaled position \(\tilde{x}\in[0,K-1]\) where K is the number of knots:
\(\tilde{x}=clip(\frac{x+1}{2}, 0, 0.9999)*(K-1)\).\\
Let \(j = \lfloor \tilde{x} \rfloor\) and \(t = \tilde{x} - j\). The linear spline is:\\
\(f_{lin}*(x) = (1-t)\theta_j*+t\theta_{j+1}\)
where \(\theta_j\) are the learnable knot values.  
Using the same scaled position \(\tilde{x}\), index j and fraction t, we compute the Catmull-Rom style tangents:
\[m_j = \begin{cases} \theta_{j+1} - \theta_j & \text{if } j = 0 \\
                      \frac{\theta_{j+1} - \theta_{j-1}}{2} & \text{otherwise} \end{cases}\]
\[m_{j+1} = \begin{cases} \theta_{j+1} - \theta_j & \text{if } j + 1 = K - 1 \\
                      \frac{\theta_{j+2} - \theta_{j}}{2} & \text{otherwise} \end{cases}\]
Using those we compute the cubic Hermite basis polynomials:
\begin{align*}
f_{cb}(x)=(2t^3-3t^2+1)\theta_j+(t^3-2t^2+t)m_j+\\
(-2t^3+3t^2)\theta_{j+1}+(t^3-t^2)m_{j+1}
\end{align*}

\subsubsection{Composite layer output}
For input vector \(\boldsymbol{x}\in\mathbb{R}^N\) the spline layer output is the concatenation of:\\
Univariate terms (for each feature i and resolution r):
\[\phi_i^{raw} = w_i^{raw} * x_i\]
\[\phi_{i,r}^{lin} = w_{i,r}^{lin} * f_{i,r}^{lin}*(x_i)\]
\[\phi_{i,r}^{cb} = w_{i,r}^{cb} * f_{i,r}^{cb}*(x_i)\]
Bivariate terms (for each pair \(i < j\) and resolution r):
\[\psi_{i,j}^{raw} = w_{i,j}^{raw} * x_i* x_j\]
\[\psi_{i,j, r}^{lin} = w_{i,j,r}^{lin} * f_{i,r}^{lin}*(x_i)* f_{j,r}^{lin}*(x_j)\]
\[\psi_{i,j, r}^{cb} = w_{i,j,r}^{cb} * f_{i,r}^{cb}*(x_i) * f_{j,r}^{cb}*(x_j)\]

All \(\phi\) and \(\psi\) terms are concatenated together for the layer's output:
\[S = \underset{i=1, r=1}{\stackrel{N,R}{\bigoplus}} [\phi_i^{raw} \|\phi_{i,r}^{lin}\|\phi_{i,r}^{cb} \| (\underset{i<j}{\stackrel{N}{\bigoplus}} [\psi_{i,j}^{raw}\|\psi_{i,j,r}^{lin}\|\psi_{i,j,r}^{cb}])]\] 
where \(\bigoplus \text{ and } \|\) denote concatenation.
For a visualisation of the spline composition method see Figure \ref{fig:spline_generation}.\\

\subsubsection{Loss function and regularisation}
The Spline layer adds to the loss of the learning algorithms as:\\
\(\mathcal{L}_{mrcps} = \mathcal{L}_{L1} + \mathcal{L}_{smooth}\).\\
\(\mathcal{L}_{L1}\) is an L1 sparsity penalty:
\[\mathcal{L}_{L1} = \lambda_{1}(\sum_{r,i,type} \|\theta_{i,r}^{type}\|_1 + \sum_{r,i,type} \|w_{i,r}^{type}\|_1)\]
where \(type\) can be \(raw\), \(lin\) or \(cb\).
\(\mathcal{L}_{smooth}\) is the smoothness (roughness) penalty:
\[\mathcal{L}_{smooth} = \frac{1}{|s| + \epsilon}\sum_{i,r, type}\sum_{k=0}^{K_r-3}(\theta_{k+2} - 2\theta_{k+1} + \theta_k)^2\]
following the P-Spline formulation \cite{eilersFlexibleSmoothingBsplines1996}. In our case the model learns the regularisation parameter \({\lambda}_1\) through backpropagation instead of choosing it via Generalised Cross Validation (GCV) while we fix the \(\frac{1}{|s| + \epsilon}\) to 0.

\subsubsection{Extracting splines from the trained model}
After a model is trained the splines it uses for inference are parametrised through the \(\theta \text{s and } w\)s. In order to explore what a model has learned and how the splines it generates can provide interpretable results, we can combine the appropriate \(\theta \text{s and } w\)s to reconstruct and visualise the learned splines. We reconstruct the full set of splines the model generates, as described above. These by themselves though, would not be a very useful visualisation and interpretability tool.  We also linearly combine all the resolutions for all spline types (linear and cubic) for each feature separately. This way we construct N splines, each the linear combination of six univariate resolutions - three linear and three cubic plus the raw spline \(\phi^{raw}(x_i) = w_i^{raw}x_i\) for feature ${x_i}$. For the arity 2 spline we construct N-1 bivariate splines each combining the feature with each one of all the other features minus itself (\(x_i, x_j \text{ for } i\neq j\)). Again for the bivariate case each spline is the linear composition of the corresponding bivariate resolutions for the raw, linear and cubic types (see Fig. \ref{fig:spline_generation} for a full description of how spline components are combined).

\subsection{Full network structure and training}
\label{methods_network}
\subsubsection{Input}
Our model is designed to deal with tabular datasets where each feature can be either numerical or categorical (treating ordinal as numerical). Numerical inputs are scaled between -1 and 1 before being passed on to the Spline layer.
Since our model is embedded in the TALENT benchmark \cite{liuTalentTabularAnalytics} we deal with categorical data using TALENT's infrastructure. Specifically TALENT by default uses ordinal encoding, assigning each category a number ranging from -1 to 1 so that all categories are equally spread over this range. This is not optimal and there are better ways to deal with categorical data (like appropriate embeddings) but we wanted to be able to directly compare our results with the ones provided by TALENT for all other models so we did not modify the default categorical data transformation.

\subsubsection{Output}
The tabular data we deal with can be either a regression problem (continuous output), a binary classification problem (binary output) or a multi classification one (categorical output with more than two categories). In the regression case the continuous output of our Spline layer is passed to a Linear layer with a single output and in the two classification cases the Spline's output is passed to a Linear layer with number of output neurons the number of classes. Their output is then treated as probabilities and passed to a Softmax layer to pick the highest probability class. In the case of regression, once the data pass through the Spline Layer and the Linear Layer for N epochs and their learning finishes then the output of the Spline Layer is passed through a Ridge Cross Validation (RidgeCV) Layer which selects the best regularization parameter for the Linear Layer's weights.

\subsection{TALENT datasets comparisons}
\label{methods_talent}
In order to compare our algorithm with the ones offered by the TALENT benchmark we created two metrics, a ranking order and a score measuring proportional difference to the MLP algorithm. The score we used for both metrics was either the \(R^2\) for regression datasets or the Accuracy (Acc) for multi or binary classification ones. We run our method, ${CoGA^2M}$, with 15 (TALENT default) repetitions. The results over the repetitions were in all datasets diverging to less than 0.1\% (data not shown). We did no hyperparameter search and run the model with a batch size of 16, a learning rate of 0.001 and with arity 2. We set it to run a maximum of 5000 epochs but with early stopping if the validation score (validation set was 10\% of the data) stopped getting better for at least 100 epochs.\\
For ranking we used the Placket-Luse model \cite{maystreFastAccurateInference2015}. The difference to the MLP results were calculated as relative difference between scores:
\[\frac{{score}_{CoGA^2M} - {score}_{MLP}}{{score}_{MLP}}\]
For the regression we used the \(R^2\) as the score while for the classifications the model's accuracy.

\subsection{Generation of OOD datasets}
\label{methods_ood_gen}
We use only a subset of the 300 datasets in the TALENT benchmark to generate OOD datasets. The first selection strategy was to use only regression datasets that had only numerical inputs. Using only numerical inputs datasets both alleviated the issue of how exactly to transform the categorical data for our ${CoGA^2M}$ algorithm and allowed us to create an OOD data generation process that did not have to deal with categorical data. That resulted in a subset of 48 datasets. The second selection had to do with the time each of our OOD generation processes (we used two) took to break up the data. For the first process we managed to create 31 OOD datasets while with the second 39. Those were partially overlapping (29 in common) subsets of the original 48 datasets. The rest of the datasets were too big (in number of samples and features) to be processed within a reasonable time frame.\\
We focused on creating OOD datasets with diversity shifts. In order to achieve this we needed to generate new datasets with data from a compact, connected subset of the interior (a region) of the dataset's convex hull assigned to the test data while the rest (100\% of the region's complement) is assigned to training (90\%) and validation data (10\%). But this region needed to be a feasible region in the sense that the data in it and the data in its complement needed to be drawn from different distributions. Since we could not directly infer the distributions of the newly generated train and test sets we used as a proxy for OOD the drop in the ability of a model to predict the OOD test data when trained on the OOD train set compared to its ability to predict the TALENT ID test data when trained on the ID train set. \\
In order to create new data cuts where a model would fail to learn the test data we created two versions of an adversarial algorithm. Each had an evolutionary process optimising the feasible region to make the test data in it as hard to predict from the train data in the complement of the feasible region. At the same time, a model learned the newly cut train data trying to maximize its result on the new test data. The two versions of the adversarial algorithm differed by the model learning on the train data. In the first case we used a K Nearest Neighbours Regressor (kNNR) with a hyperparameter search over its distance metric (Manhattan and Euclidean), its weights distribution (uniform and distance) and its k (1, 3, 5, 7, 9 and 11). In the second case we used \(\nu\) Support Vector machine Regression (\(\nu\)SVR) with \(\nu = 5\).\\
For both versions, the OOD generating process was the Covariance Matrix Adaptation (CMA) evolutionary strategy \cite{hansenCMAEvolutionStrategy2023} with a population size of 20. For both the kNNR and the \(\nu\)SVR models and for each run of the algorithm we followed the decreasing \(R^2\) of the models' results on the CMA updating train, validation and test datasets. For every drop in the \(R^2\) we saved that train/val/test data cut. We asked the algorithms to stop at either after 20 such cuts or at 300 epochs (whichever came first). That produced a dataset specific number of different cuts, each one progressively harder to be learned by the kNNR or \(\nu\)SVR models. We run each algorithm on each dataset 5 times concatenating the cuts over all runs and sorting them according to descending \(R^2\). 

\subsection{OOD datasets comparisons}
\label{methods_ood_comp}
Once we had OOD cut data for as many datasets as possible from the kNNR (31 datasets) and the \(\nu\)SVR (39 datasets) based processes we compiled a list of datasets with their cuts by choosing the results from one of the two cutting algorithms for each dataset. The selection process was: 
\begin{enumerate}
    \item If an algorithm had not produced an \(R^2\) in the last (smallest \(R^2\)) cut smaller than 0 then do not consider this algorithm/dataset combination a valid series of cuts. 
    \item If only one of the two algorithms had produced a valid cuts series, use this.
    \item If both algorithms had produced valid cuts then use the one from the kNNR unless
    \item the kNNR's maximum \(R^2\) (from the first cut of the series) was smaller than 0 and the \(\nu\)SVR's first cut \(R^2\) was larger than 0, in which case use the \(\nu\)SVR.
\end{enumerate}
The logic behind the 3rd rule was to keep as many of the kNNR cuts as possible since we deemed this to be the algorithm that would bias the follow up comparison between ${CoGA^2M}$ and LightGBM the least. The logic behind the 4th rule was that if the kNNR algorithm at its first cut (before the evolution algorithm had time to select appropriate regions) had failed to learn the data then it was not providing the CMA a good enough signal for appropriate region selection and then the \(\nu\)SVR algorithm was to be preferred. The final list consisted of 40 datasets (out of the 48 initial ones), 28 were cut with the kNNR and 12 with the \(\nu\)SVR algorithm.\\
For each dataset and the selected cutting model we grouped the sorted series of cuts into 3 sets, each one having cuts with \(R^2\) in the top, middle and bottom quantiles of the \(R^2\) range. For each (ID) dataset we defined its OOD dataset as the group of all the cuts of the bottom quantile (the ones with the worst \(R^2\)s). The number of such cuts for each dataset finally used for our comparisons is shown in Table \ref{table:differences} in the column N.\\
We run the ${CoGA^2M}$ and the LightGBM models (to compare with a model that does really well on the regression part of the benchmark) on both the ID and the OOD datasets. We then used the generated \(R^2\) to exclude some datasets from the final comparison. We would exclude a dataset using the following rules:
\begin{enumerate}
    \item If both the LightGBM and ${CoGA^2M}$ algorithms did not achieve an \(R^2\) larger than 0 in the OOD data. In this case both algorithms were considered to fail to learn the OOD data. There were 13 datasets like that.
    \item If both algorithms achieved an \(R^2\) larger than 0.9 in the OOD data. In this case both algorithms are considered to succeed in the OOD data or (most probably) our OOD generation method did not manage to break the dataset appropriately to create an OOD test set. There were 3 datasets in this group.
    \item If both algorithms fail to achieve an \(R^2\) larger than 0.5 in the ID data. In this case neither algorithm can learn the ID dataset, so there is no point in comparing how well they managed to learn the OOD one. There are 6 datasets in this group 2 of which also exist in the \(r_{OOD}^2 < 0\) set.
\end{enumerate}
This left 22 datasets on which we compared ${CoGA^2M}$ and LightGBM. For each dataset we first calculated the differences between the \(R^2\)s of the two algorithms on the ID and the OOD data \[\Delta_{CoGA^2M} =  r_{CoGA^2M, OOD}^2  - r_{CoGA^2M, ID}^2\] and 
\[\Delta_{LightGBM} =  r_{LightGBM, OOD}^2  -\\ r_{LightGBM,ID}^2\]
and finally the difference of the differences
\[\Delta\Delta = \Delta_{CoGA^2M} - \Delta_{LightGBM}\]
We then tested whether the \(\Delta\Delta_{dataset}\) series across the 22 datasets is significantly larger than 0. We did this using both a t-test (one-tailed) and bootstrapping with 10000 draws (in order to mitigate the small number of samples).

\section{Results}
\label{results}
\subsection{Spline visualisation and interpretability}
\label{spline_visualisation}

\input{Fig3}

${GAMs}$ and ${GA^2Ms}$ were designed to provide interpretable results. As the models evolved in complexity, significant effort was aimed into keeping the resulting splines as interpretable as possible. From the initial ${GAM}$ idea \cite{hastieGeneralizedAdditiveModels} with its local scoring algorithm to automatically calculate the smallest appropriate number of knots, to the Penalised-Splines ${GAMs}$ \cite{eilersFlexibleSmoothingBsplines1996}, for more efficient knot calculation, to the higher order arity ${GA^2Ms}$ \cite{louAccurateIntelligibleModels2013} with the FAST measure to pick the most appropriate feature combinations and the ${CGA^2M+}$s \cite{watanabeConstrainedGeneralizedAdditive2021} enforcing monotonicity on certain features, interpretability has been at the core of ${GAM}$ development. Here we show that our multiple resolution, P-Spline based ${GA^2M}$ without monotonicity or any separate (outside the backpropagation algorithm) method for feature selection provides interpretable results. Figures \ref{fig:arity2_splines} and S3 (Supplementary Materials) show the splines learned by our ${CoGA^2M}$ model on four datasets from the TALENT benchmark. The airfoil\_self\_noise succeeds at learning the data with an ${R^2}$ on the test set of 0.92 \(\pm\) 0.0038, the concrete\_compression\_strength with an ${R^2}$ of 0.934 \(\pm\) 0.0063 and the stock with ${R^2}$ of 0.985 \(\pm\) 0.0004. We also show the splines on the fifa dataset where it achieves a lower ${R^2}$ of 0.651 \(\pm\) 0.0008. In all four cases the arity 1 and arity 2 plots show trends that make sense for the features they represent. We have found that interesting insights can be drawn from visually exploring not just the splines themselves but in the case of the 1d splines also the individual resolution sub-splines that compose them. For the case of the airfoil\_self\_noise and the fifa datasets we show in Figs. S1 and S2 (Supplementary Materials) all the sub-splines that comprise the final 1d splines of all the features for each dataset.

An important point to note in the graphs, is the fact that large regions of input space are not supported by data, especially in the arity 2 splines. In some cases, the splines seem to interpolate logically between data rich regions (e.g. the length and velocity 1d splines of the airfoil\_self\_noise). In other cases (like the release\_clause\_eur x height\_cm 2d spline of the fifa dataset in Fig. \ref{fig:arity2_splines} or the Superplasticizer 1d spline of the concrete\_compressive\_strength in Fig. S3 the splines show structure (large first and second derivatives) in regions where there are no data to support those. Whether this structure is a failure of spline based methods or an often appropriate extrapolation that only such methods can achieve was the question we tried to answer by breaking the datasets into OOD train and test sets (see \ref{ood_benchmark}). 

\subsection{TALENT benchmark}
The TALENT benchmark comes with 300 datasets divided in three categories, regression sets (100 datasets), binary classification sets (120 datasets) and full classification ones (80 datasets). It also comes with the results on all of these datasets for 29 models for regression and 31 for classification. We run the ${CoGA^2M}$ model on all of the datasets and then used the Packett-Luce metric to rank all model results. ${CoGA^2M}$ achieves an 11th/29 place in the regression set and 7th/32 places in the binary and the full classification sets (see Figure \ref{fig:packett_luce_ranking}). These results put the ${CoGA^2M}$ model in the top third of all tested models in the benchmark.
\input{Fig4}
To give a more comprehensive understanding of how the models do over the range of datasets tested we also calculated the percentage change that each model achieved on each dataset compared to the MLP model. In Figure \ref{fig:mlp_ratios} we show over all models the box and whiskers plot with the box showing the quartiles of the results distribution for this model over the category's datasets, the whiskers extend to 1.5 * InterQuartile Distance (IQR) and the circles are the results that lie beyond this range.  As can be seen by the figure, ${CoGA^2M}$ does on average just better than the MLP but with its distribution shifted to the positive (shown by the larger positive box and more positive individual datasets beyond the whisker range). This behaviour is comparable to the results produced by the best models in the datasets (e.g. tabr, realmlp, catboost, LightGBM, xgboost, modernNCA).

\input{Fig5}

\subsection{OOD benchmark}
\label{ood_benchmark}
As described in \ref{methods_ood_comp} we used the ${CoGA^2M}$ and the LightGBM models' \(R^2\) results from both the ID and the OOD datasets to calculate the ID - OOD differences for each model and each dataset that was selected to take part in the final comparison. We wanted to ask whether, for a given dataset, the ${CoGA^2M}$ model would be expected to show a smaller drop in the OOD average result vs its ID result compared to the equivalent drop in the results of the LightGBM model. To do this we first calculated for each dataset the model's difference between the ID and the average OOD \(R^2\)s \(\Delta_{CoGA^2M}\) and \(\Delta_{LightGBM}\) and then the between model difference of these differences \(\Delta\Delta\) (as described in \ref{methods_ood_comp}). In Table \ref{table:differences} we show these calculated differences and differences of differences (\(\Delta\Delta\)) for all 22 datasets that we use to compare the two models. We show the actual \(R^2\) values for the ID and OOD datasets, as well as which method was used to generate the OOD data, for all 40 models in Table S1 (Supplementary Materials).\\

\begin{table}[h!]
\begin{tabular}{ |p{2.3cm}||p{1.3cm}|p{1.2cm}|p{1.35cm}|p{0.6cm}| }
 \hline
 \multicolumn{5}{|c|}{Models' \(R^2\) differences and number of OOD cuts} \\
 \hline
 Dataset& ${CoGA^2M}$ OOD - ID& LGBM OOD - ID &\(\Delta_{CoGA^2M} - \Delta_{LGBM}\)&N\\
  \hline
3D-Estimation-using-RSSI-of-WLAN-dataset-complete-1-target	&-0.164	&-0.075	&-0.089	&13\\
Another-Dataset-on-used-Fiat-500-(1538-rows)	&-0.137	&-0.017	&-0.120	&12\\
Bias-correction-r	&-0.872	&-0.198	&-0.674	&34\\
Brazilian-houses-reproduced	&-0.046	&-0.164	&\textbf{0.118	}&11\\
IEEE80211aa-GATS	&-0.397	&-0.470	&\textbf{0.073}	&49\\
Laptop-Prices-Dataset	&-0.619	&-2.736	&\textbf{2.117}	&3\\
MiamiHousing2016	&-0.755	&-1.219	&\textbf{0.464}	&16\\
airfoil-self-noise	&-0.197	&-0.328	&\textbf{0.131}	&4\\
bank8FM	&-0.245	&-0.869	&\textbf{0.623}	&24\\
concrete-compressive-strength	&-0.193	&-0.455	&\textbf{0.262}	&19\\
dataset-sales	&-0.264	&-0.312	&\textbf{0.048}	&1\\
delta-elevators	&-0.017	&-0.062	&\textbf{0.045}	&9\\
fifa	&-0.324	&-0.458	&\textbf{0.134}	&5\\
house-sales-reduced	&-0.708	&-1.147	&\textbf{0.439}	&26\\
houses	&-0.245	&-0.269	&\textbf{0.024	}&53\\
kin8nm	&-0.506	&-0.360	&-0.146	&15\\
pole	&-0.106	&-0.126	&\textbf{0.020}	&53\\
puma8NH	&-0.045	&-0.115	&\textbf{0.070}	&9\\
qsar-aquatic-toxicity	&-0.220	&-0.429	&\textbf{0.208}	&1\\
stock	&-0.162	&-0.470	&\textbf{0.308	}&12\\
treasury	&-0.097	&-1.346	&\textbf{1.249	}&19\\
wind	&-0.244	&-0.370	&\textbf{0.126	}&10\\
 \hline
\end{tabular}
\\
\caption{Differences of \(R^2\) between ID and OOD data for each dataset for both the ${CoGA^2M}$ and the LightGBM models and the differences of these differences between the two models. With bold we showcase the differences of the differences where ${CoGA^2M}$ does better than LightGBM.}
\label{table:differences}
\end{table}
We then tested whether these \(\Delta\Delta\)s were expected to be drawn from a distribution with an expected value larger than 0. To do this we performed a one-tailed t-test and a bootstrap non-parametric test with 10000 iterations. In the t-test the t-statistic was 2.118 and the one-tailed p-value 0.0232, above the 0.05 value which would indicate that our results were drawn from a larger than 0 mean distribution by chance with a chance higher than 5\%. The bootstrap results over 10000 iterations create an observed mean of 0.247 with a one-tailed p-value of 0.006 and 95\% confidence intervals of 0.0481 and 0.4877. In Figure \ref{fig:DD_hist} we show the distribution of the means of the 10000 bootstrap generated \(\Delta\Delta\) sets.

\input{Fig6}

\section{Conclusion}
We frame OOD as a statistical problem to develop an intuition on why ${CoGA^2M}$ is, in many OOD scenarios, an appropriate model. Denote a dataset as a matrix \(O =(Y,X)\) with sample data inputs X and outputs Y where \(Y = f(X) + \epsilon\). We wish to learn \(\{f(x):x\in\Omega_0\}\) over the population domain \(\Omega_0\) from sampled data over a restricted domain \(\Omega_s \subseteq \Omega_0\). The $n$ rows of \(O\) are taken to be \textit{i.i.d} realisations from the probability distribution of random vector \((y_i,x_i)\) with density \(f_{Y,X}\). For finite samples, it is always the case that \(\Omega_s \neq \Omega_0\) and for ID scenarios \(\Omega_s \rightarrow \Omega_0 \text{ as sample size } n \rightarrow \infty\). In the OOD case, the sampling process induces a distribution with density \(f_s \neq f_{Y,X}\)  arising from a sampling processes \(R(y,x)\) with one of two possible special characteristics. In the first case \(R = I(y,x \in \Omega_1)\) for some \(\Omega_1 \subset \Omega_0\), so that the sampled domain is restricted to a strict subset of the population domain, and \(\Omega_s \rightarrow \Omega_1 \text{ as } n \rightarrow \infty\). In the second case, \({\rm supp}(f_s)={\rm supp}(f_{Y,X})\) so that \(\Omega_s \rightarrow \Omega_0\), but \(f_s(y,x)\neq f_{Y,X}(y,x)\) for some \((y,x)\in \Omega_0\). That is, the sampling process covers the full domain but with a distorted distribution. Importantly, there exist regions \(\mathcal{R} \subseteq \Omega_0\) of high utility \(U(y,x)\) for which \(f_s(y,x) \ll f_{Y,X}(y,x)\), so that these regions are under-represented in the sampled data relative to their importance.

Let the learner define (potentially infinite-dimensional) model space \(F\) with respect to loss function \(\mathcal{L}\), and define the sample loss for choice \(f\in F\) by \(\mathcal{L}(Y,X\in \Omega_s;f)\). Then 
\[\hat{f} = \underset{(f \in F)}{\rm argmin}~ \mathcal{L}(Y,X \in \Omega_s,f)\]
is learnt from the sample data satisfying \(\hat{f} \rightarrow f_1 \text{ as } n \rightarrow \infty\), where
\[f_1 = \underset{(f \in F)}{\rm argmin}~ E[\mathcal{L}(y,x \in \Omega_1,f)].\]
If \(f_1=f_0\) such that
\[f_0 = \underset{(f \in F)}{\rm argmin}~ E[\mathcal{L}(y,x \in \Omega_0,f)]\]
then all OOD predictions will be optimally accurate.
Arguments from the statistical literature on incomplete/coarse data give
\[E[\mathcal{L}(y,x \in \Omega_1,f)]= \iint \pi(y,x)\mathcal{L}(y,x \in \Omega_0,f)dv(y,x),\]
where \(v(y,x)\) is a dominating probability measure and \(\pi(y,x) = Pr(R=1\mid y,x) \in [0,1]\) can take deterministic values 1 or 0 (i.e. an individual is either selected or not selected) - case 1 - or even values in (0,1) (some individuals in the range are probabilistically selected) - case 2.  
However, it cannot be guaranteed that \(f_0\) is a solution to the above problem because the range of Y could be restricted or, even more severely, the codomain and domain could be jointly constrained. Correlation shifts occur when this constrained varies between the train and test data sets (see Fig. \ref{fig:id_ood_schematic}(c)). In this work we concentrate on datasets that do not show such correlation shifts, hence, for our cases, the following assumption is required:

\textbf{Assumption 1}: The sample range only depends on values of the inputs, and does not depend on the values of the output (either purposefully or inadvertently), such that \(\pi(y,x)=\pi(x)\). Then
\[E[\mathcal{L}(y,x \in \Omega_1,f)]= \int \pi(x) \int \mathcal{L}(y,x \in \Omega_0,f)dv(y,x), \]

\noindent which is a linear combination of input-stratified optimisation problems with no constraints on the codomain and, hence, one that is minimised by \(f=f_0\).  We note that it can also be minimised by more complex \(f^*\) empirically indistinguishable from \(f\) over \(\Omega_0\) when the population domain \(\Omega_0\) is a strict subset of the entire domain of \(f\), but presume that model selection will effectively avoid choosing \(f^*\) over \(f\). However, a further assumption is still needed:

\textbf{Assumption 2}: \(\pi(x)\) induces ‘sufficient variation’ in \(\Omega_1\) such that there are no simpler \(g \in F\) satisfying \(E[\mathcal{L}(y,x \in \Omega_1;g)]=E[\mathcal{L}(y,x \in \Omega_1;f_0 )]\) e.g. cases where \(g(x)=f_0 (x)\) for \(x \in \Omega_1\) but \(g(x)\neq f_0 (x) \text{ for } x \in \Omega_0\setminus \Omega_1\) (\(g\) is simpler in the sense that model selection would almost surely choose \(g\) over \(f_0\) as \(n \rightarrow \infty\)).

Assumption 2 is required to prevent dangers-of-extrapolation cases where \(f\) is highly non-linear over \(\Omega_0\) but \(\Omega_1\) contains only regions (possibly narrow ones) over which the behaviour of \(f\) is empirically indistinguishable from e.g. linear \(g(x)\). \
The \(CoGA^2M\) setup has the advantage that Assumption 2 is more plausible in the lower (one or two)-dimensional domains of each compositor function than it is for the entire multi-dimensional support of \(f\). So compared with a full-domain alternative, in case 1, \(CoGA^2M\) induces an \(\Omega_1\) closer to \(\Omega_0\). In case 2, the distributional distortion induced by \(\pi(x)\) is less severe when projected onto the low-dimensional domain of each compositor function, so the effective sampling density within each compositor domain is closer to the population density. 

Several lessons emerge from this work. First, spline-based ${GA^2M}$ models can compete with state-of-the-art models on standard tabular benchmarks without sacrificing interpretability, provided they are equipped with sufficient expressiveness through multi-arity basis functions. Second, the smooth inductive bias inherent in spline methods appears to be not merely an aesthetic or interpretability choice but a functional one: it offers a measurable advantage under diversity shift, where models must generalise to unseen regions of input space. Third, our adversarial OOD generation procedure revealed that many datasets (12 out of 40) commonly used in tabular benchmarks are surprisingly resistant to diversity shift - both models continued to perform well (with OOD ${R^2}$s larger than 0.5) - suggesting that standard benchmark performance may overstate real-world robustness. Finally, we found that the choice of proxy learner in the adversarial loop (kNNR vs. $\nu$SVR) materially affects which datasets can be successfully broken, highlighting that OOD benchmark construction is itself a non-trivial design decision that deserves more attention from the community.

\section{Future work}
Several questions have been raised by this work. First, we did not fully explore the importance of the multi-resolution character of our model. Initial experiments showed that breaking up the splines into separate, multi-knot resolution, linear and cubic parts and allowing the backpropagation algorithm to learn their additive weights (with an L1 regularisation) improved performance. This claim though will need to be adequately substantiated in future work with appropriate comparisons between our and GAM based models that do not decompose splines. Second, our ${CoGA^2M}$ model was run with fixed hyperparameters and a single repetition, without the extensive tuning afforded to other models in the TALENT benchmark; a systematic hyperparameter search — particularly over the set of resolutions, the arity, and the learning rate — could reveal further performance gains. Third, our treatment of categorical features relied on TALENT's simple linear encoding rather than learned embeddings, which likely underestimates the model's potential on categorically rich datasets. Initial efforts with a one-hot encoding faced computability problems - mainly due to the exponential growth of the arity 2 components - so efforts to create a more nuanced encoding scheme of the categorical features could potentially increase the model's results in such datasets. Fourth, our OOD generation procedure was restricted to regression datasets with purely numerical features, leaving open the question of how spline-based models behave under distribution shift in classification tasks and in the presence of categorical variables. Additionally, our OOD generation could be improved by more direct OOD metrics compared to the proxy ability of a model to solve the test data. Finally, an important extension of our work would be to generate datasets not only with diversity but also correlation shifts. The correlation shift setting — where spurious feature–target associations change between training and test — remains unexplored and would require careful experimental design to disentangle genuine from spurious relationships. Finally we concentrate only on diversity shifts, leaving the correlation shift scenario for future work, as it would require intensive hyperparameter optimisation to appropriately disentangle spurious from genuine feature–target relationships.

\section{Acknowledgments}
GD and SS acknowledge the support of Engineering and Physical Sciences Research Council (EPSRC) grant "UKRI1076 Element-compositor methods for out-of-distribution machine learning".\\
PC was partly funded by UK Economic and Social Research Council award ES/S012486/1 (MiSoC).\\
DM acknowledges the support of the UKRI AI programme, and the Engineering and Physical Sciences Research Council, for CHAI - Causality in Healthcare AI Hub [grant number EP/Y028856/1].\\
For the purpose of open access, the authors have applied a Creative Commons Attribution (CC BY) licence to any Author Accepted Manuscript version arising.\\
We used Claude Opus 4.6 Extended as an editor. The whole manuscript was written manually by the authors and then given to Claude to do small changes in the format of phrases and the choice of certain words. The proposed changes were then re-reviewed by the authors.

\bibliographystyle{IEEEtran}
\bibliography{bibliography/bibliography}
\vspace{1.0cm}

\end{document}


\title{Rethinking ${ GA^2Ms}$: performance and out-of-distribution generalisation. Supplemental materials}


\maketitle
\input{SM_Fig1}
\clearpage

\input{SM_Fig2}
\clearpage

\input{SM_fig3}

\clearpage

\begin{table*}
\begin{tabular}{ |p{5cm}||p{2cm}|p{2cm}|p{2cm}|p{2cm}|p{2cm}| }
 \hline
 \multicolumn{6}{|c|}{Models' \(r^2\) for ID and OOD datasets} \\
 \hline
 Dataset& ${CoGA^2M}$ \(R_{ID}^2\)&LGBM \(R_{ID}^2\) &${CoGA^2M}$ \(R_{OOD}^2\)&LGBM \(R_{OOD}^2\)&OOD generator\\
 \hline
 3D-Estimation-using-RSSI-of-WLAN-dataset-complete-1-target&0.995    &0.999      &0.831   &0.925 &kNNR\\
 Another-Dataset-on-used-Fiat-500-(1538-rows)&0.853    &0.828      &0.730   &0.813 &kNNR\\
Bias-correction-r&0.924&0.933      &0.111   &0.748 &\(\nu\)SVR\\   
Brazilian-houses-reproduced  &0.988    &0.989       &0.944   &0.831 &kNNR\\  
IEEE80211aa-GATS   &0.994    &0.979      &0.635   &0.554 &kNNR\\ 
Laptop-Prices-Dataset &0.666&0.646 &-0.012&-2.356 &kNNR  \\ 
MiamiHousing2016 &0.933 &0.929 &0.199 &-0.256 &\(\nu\)SVR\\  
airfoil-self-noise  &0.92&0.938 &0.704&0.578 &\(\nu\)SVR\\  
bank8FM &0.963&0.956&0.706 &0.045 &kNNR\\ 
concrete-compressive-strength &0.934&0.909&0.75&0.478 &kNNR\\  
dataset-sales  &0.631&0.524&0.369&0.215 &kNNR\\   
delta-elevators &0.646&0.564&0.63&0.504 &kNNR\\  
fifa  &0.651&0.618 &0.333&0.168 &kNNR\\
house-sales-reduced &0.862 &0.825&0.061&-0.472 &\(\nu\)SVR\\  
houses &0.81&0.839&0.568&0.573 &\(\nu\)SVR\\ 
kin8nm &0.611 &0.824&0.119&0.474 &kNNR\\
pole &0.958&0.989&0.852&0.864 &\(\nu\)SVR\\
puma8NH&0.666&0.606&0.621&0.493 &kNNR\\   
qsar-aquatic-toxicity &0.526&0.482&0.252&-0.048 &kNNR\\   
stock &0.985&0.98&0.802&0.45 &kNNR\\ 
treasury  &0.995&0.992&0.904&-0.275 &kNNR\\
wind  &0.81&0.783&0.569 &0.418 &kNNR\\
 \hline
1000-Cameras-Dataset &0.571&0.208&-0.583&-1.699&kNNR  \\ 
3D-Estimation-using-RSSI-of-WLAN-dataset&0.998&0.999&-2.54&-1.687 &kNNR   \\
CookbookReviews  &0.069&-0.009&-1.22 &-1.011 &kNNR  \\ 
Parkinson-Multiple-Sound-Recording &0.096&-0.0618&-0.529&-0.575&kNNR  \\ 
Wine-Quality-red  &0.305 &0.385 &-0.303&-0.074 &kNNR  \\ 
cpu-act &0.985&0.984 &-111.127 &-5.654 &\(\nu\)SVR  \\ 
cpu-small  &0.977&0.979 &-231.819&-5.47 &\(\nu\)SVR  \\ 
debutanizer &0.757 &0.814 &-1.072 &-0.721 &kNNR  \\ 
house-16H-reg &0.587 &0.634 &-0.215 &-0.088&\(\nu\)SVR  \\ 
house-8L &0.676&0.673 &-0.81&-0.997 &\(\nu\)SVR  \\ 
space-ga &0.701&0.683 &-0.154&-0.235 &kNNR  \\ 
sulfur &0.717&0.825&-1.32 &-1.301  &kNNR  \\ 
\hline
CookbookReviews   &0.069 &-0.009&-1.22 &-1.011 &kNNR   \\ 
Goodreads-Computer-Books &0.148&-0.292&0.594 &0.51 &\(\nu\)SVR \\    
Parkinson-Multiple-Sound-Recording&0.096 &-0.061 &-0.529 &-0.575 &kNNR  \\  
Wine-Quality-red &0.305&0.385&-0.303&-0.074 &kNNR \\  
Wine-Quality-white  &0.371&0.463 &0.286 &0.311  &kNNR\\ 
wine+quality &0.304 &0.381 &0.147&0.252 &kNNR \\ 
\hline
analcatdata-supreme &0.971 &0.979&0.98&0.99 &\(\nu\)SVR \\
combined-cycle-power-plant&0.945&0.962 &0.911&0.9 &kNNR\\
weather-izmir &0.993&0.99  &0.974 &0.989  &kNNR  \\
\hline
\end{tabular}
\\
\caption{\(R^2\) of ID and OOD data for each dataset with OOD data for both the ${CoGA^2M}$ and the LightGBM models. The top sub part of the table shows the datasets that were finally used in the comparison between ${CoGA^2M}$ and LightGBM. The middle part shows the datasets that did not make it into the above list because both the ${CoGA^2M}$ and the LightGBM OOD \(R^2\)s were smaller than 0. The third part shows datasets that were removed because both models did not achieve an \(R^2\) larger than 0.5 in the ID data. Finally the last part shows datasets removed because both models managed to achieve larger than 0.9 \(R^2\)s in both the ID and OOD data.}
\label{table:all_r2_id_ood}
\end{table*}

%% file: Fig1.tex
\begin{figure}[h!]
    \centering
     \subfloat[In-distribution]{%
        \includegraphics[width=0.45\textwidth]{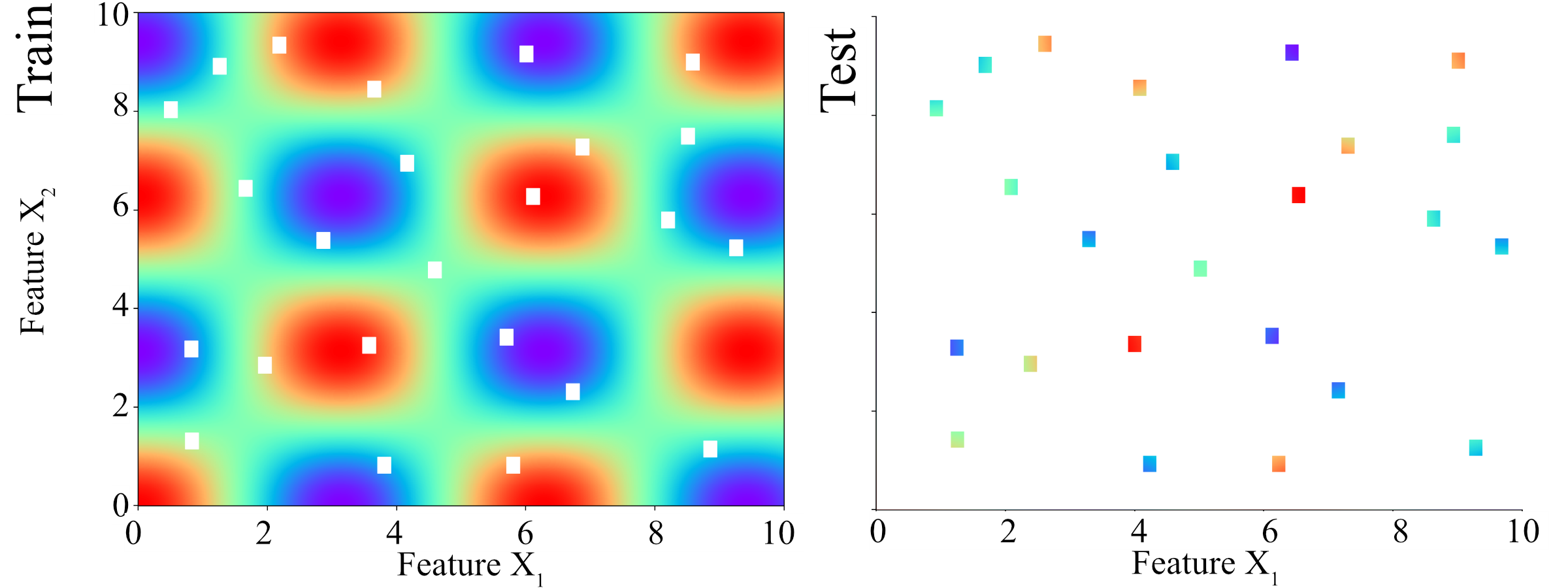}
    }\hfill
    \subfloat[Out-of-distribution diversity shift]{%
        \includegraphics[width=0.45\textwidth]{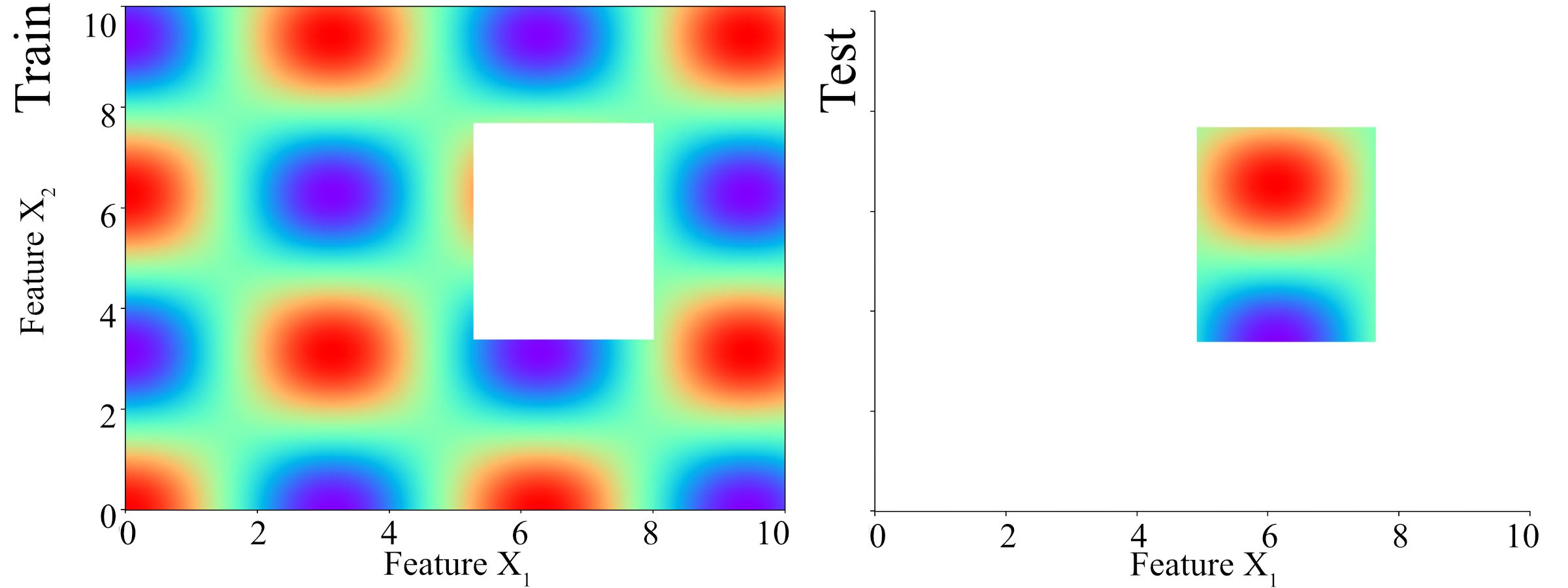}
    }\hfill
   \subfloat[Out-of-distribution correlation shift]{%
        \includegraphics[width=0.45\textwidth]{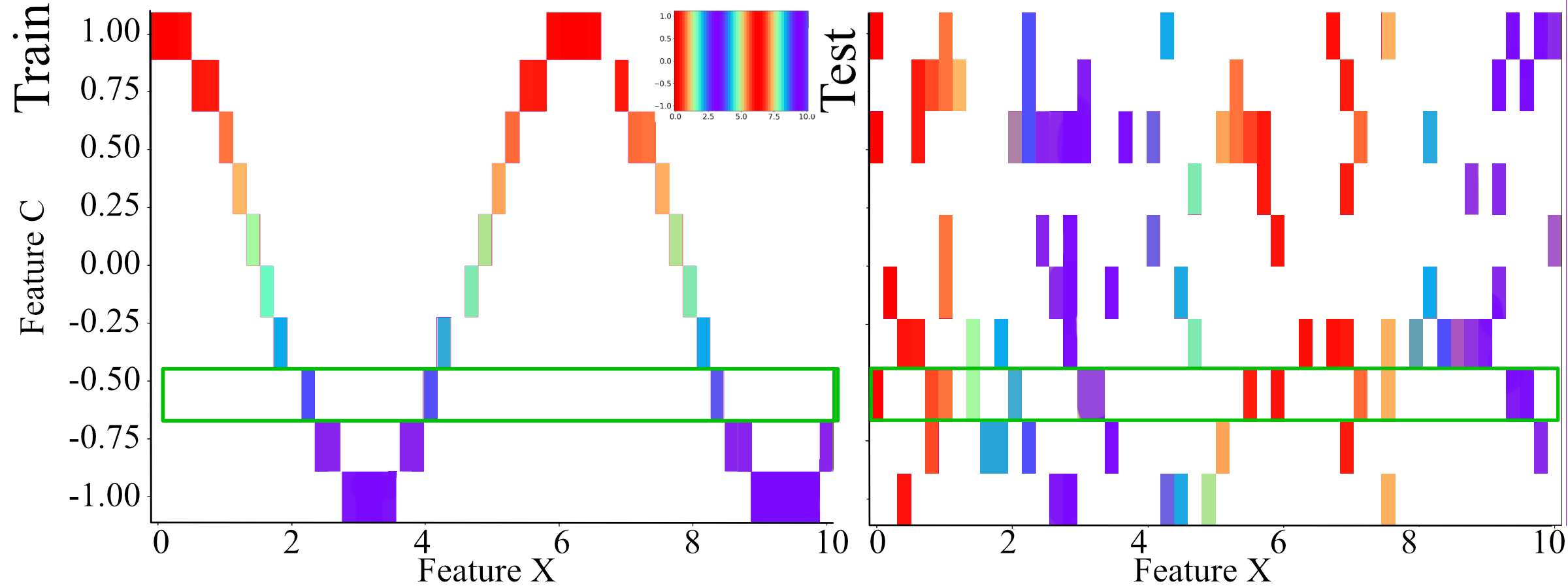}
    }\hfill
    \caption{Schematic diagram of in-distribution (a), out-of-distribution with diversity shift (b) and out-of distribution with correlation shift (c) tabular datasets. In all three sub-figures the data are 2 dimensional (2 features) and the colour indicates the Y values of the examples. The white regions indicate missing data. The left plots show the training data and the right ones the test sets. In (a) and (b) the data are generated by the function \(Y = cos(X_1) * cos(X_2)\). In (a) the test data cover small patches of the whole distribution and are uniformly distributed. The size of the missing patches is not larger than the characteristic length of the latent features of the dataset, the frequencies of the two cosines. In (b) on the other hand the train set's missing data form a simply connected region whose size is of the same order as the characteristic length of the latent feature. The test data are drawn from a distribution that is diverse (diversity shifted) from the one generating the train set. In (c) the underlying causal structure is \(Y=cos(X)\). In the left plot feature C correlates with Y (\(C=Y\)) as only data with that correlation are shown to the model (insert shows data from full range with no C, Y correlation). In the right plot (test set) that correlation is broken since for each value of C there are multiple examples with \(Y\neq C\). To make the point clear, the green rectangle shows the Y values for a specific value of C. In the train set all Y values are the same (equal to C) while in the test set Y takes a number of values uncorrelated with C. In this case the model will focus on the spurious correlation and will fail to learn the underlying structure. The schematic makes clear the hard to intuit nature of tabular correlation shifts in comparison with other types of data (e.g. images).}
    \label{fig:id_ood_schematic}
\end{figure}

%% file: Fig2.tex
\begin{figure*}
    \centering
    \begin{subfigure}[c]{0.65\textwidth}
    \centering
    \includegraphics[width=\textwidth]{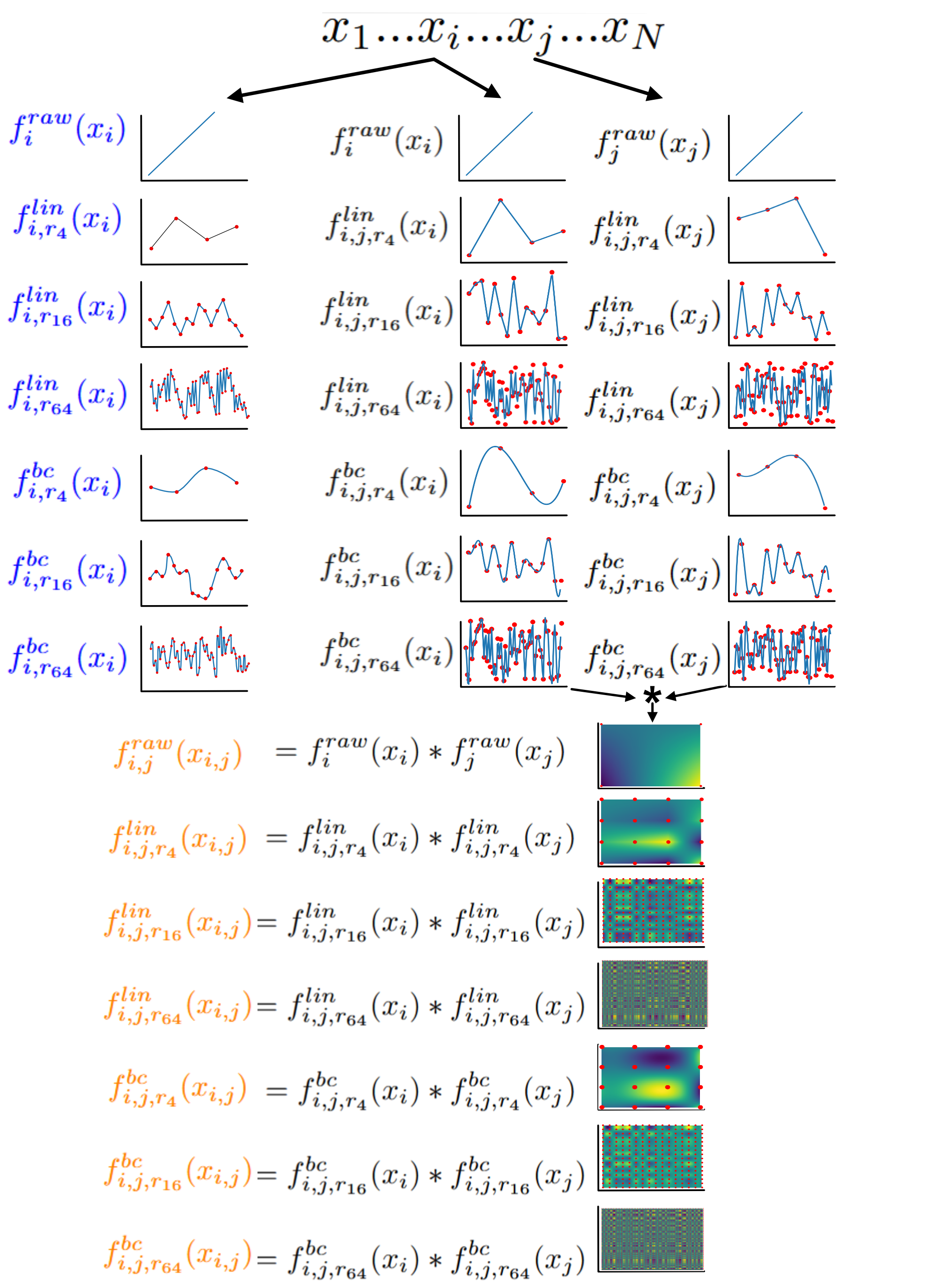}
    \caption{Break down of spline generation from sub-splines}
  \end{subfigure}
  \hspace{-3.5em}
  \begin{subfigure}[c]{0.4\textwidth}
    \centering
    \includegraphics[width=\textwidth]{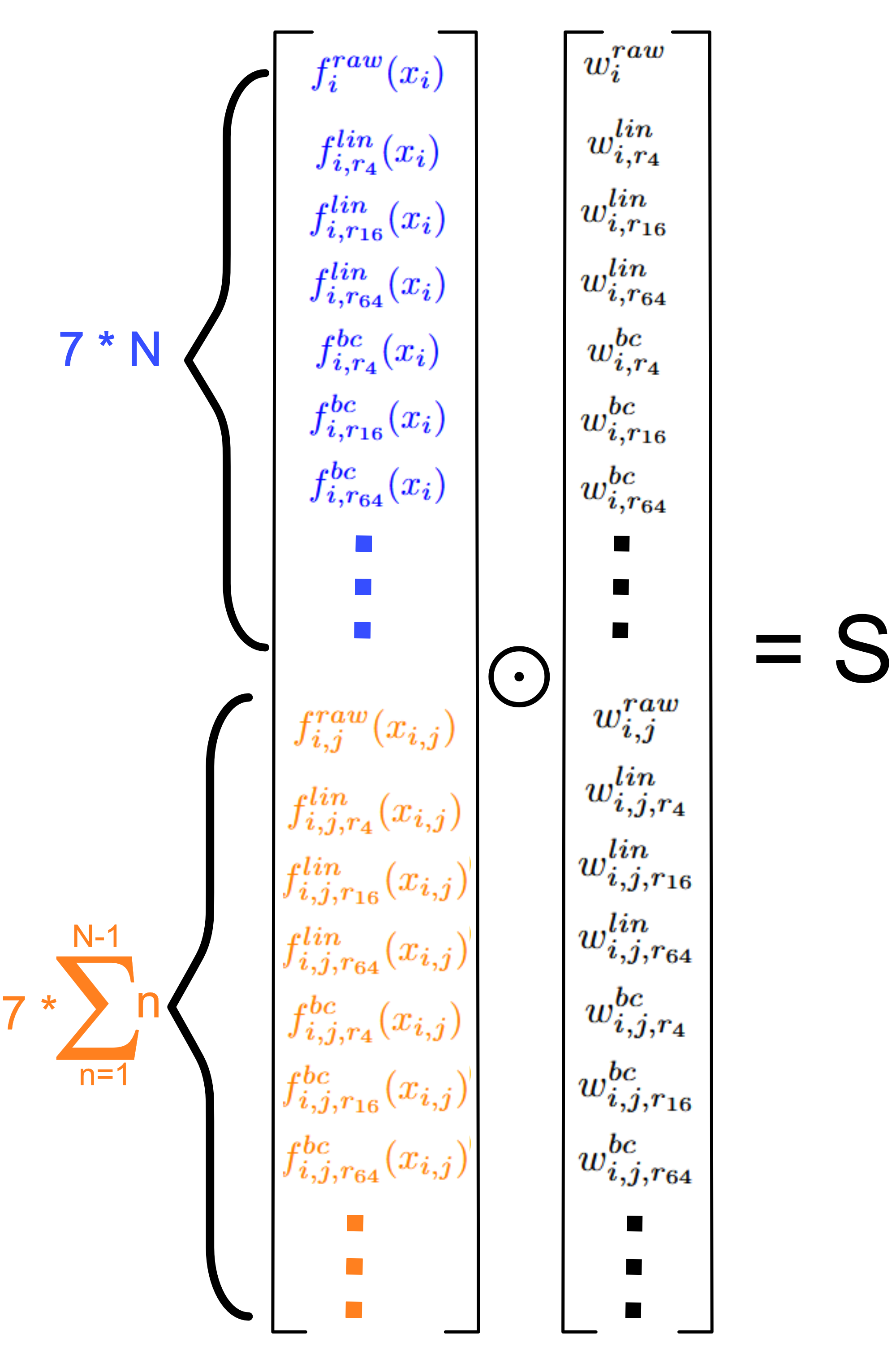}
    \caption{Spline layer output}
  \end{subfigure}
  
    \caption{Visualisation of the spline composition. (a) For each feature \(x_i\) and each of the three resolutions we use in this work (4, 16 and 64 knots) we compute the univariate and bivariate linear splines as well as the cubic Hermite ones. We also compute \(f_i^{raw}(x_i) = x_i\). For each input \(x_i\) and each resolution \(r\) the bivariate splines (linear and cubic) are the product \(f_i(x_i) * f_j(x_j)\) \(\forall j \text{ s.t. } i < j\). (a) The vector that is the result of the concatenation of all the generated splines acting on their individual features is then linearly combined to produce the layer's output.}
    \label{fig:spline_generation}
\end{figure*}

%% file: Fig3.tex
\begin{figure*}[h!]
    \centering
     \subfloat[airfoil\_self\_noise]{%
        \includegraphics[width=0.87\textwidth]{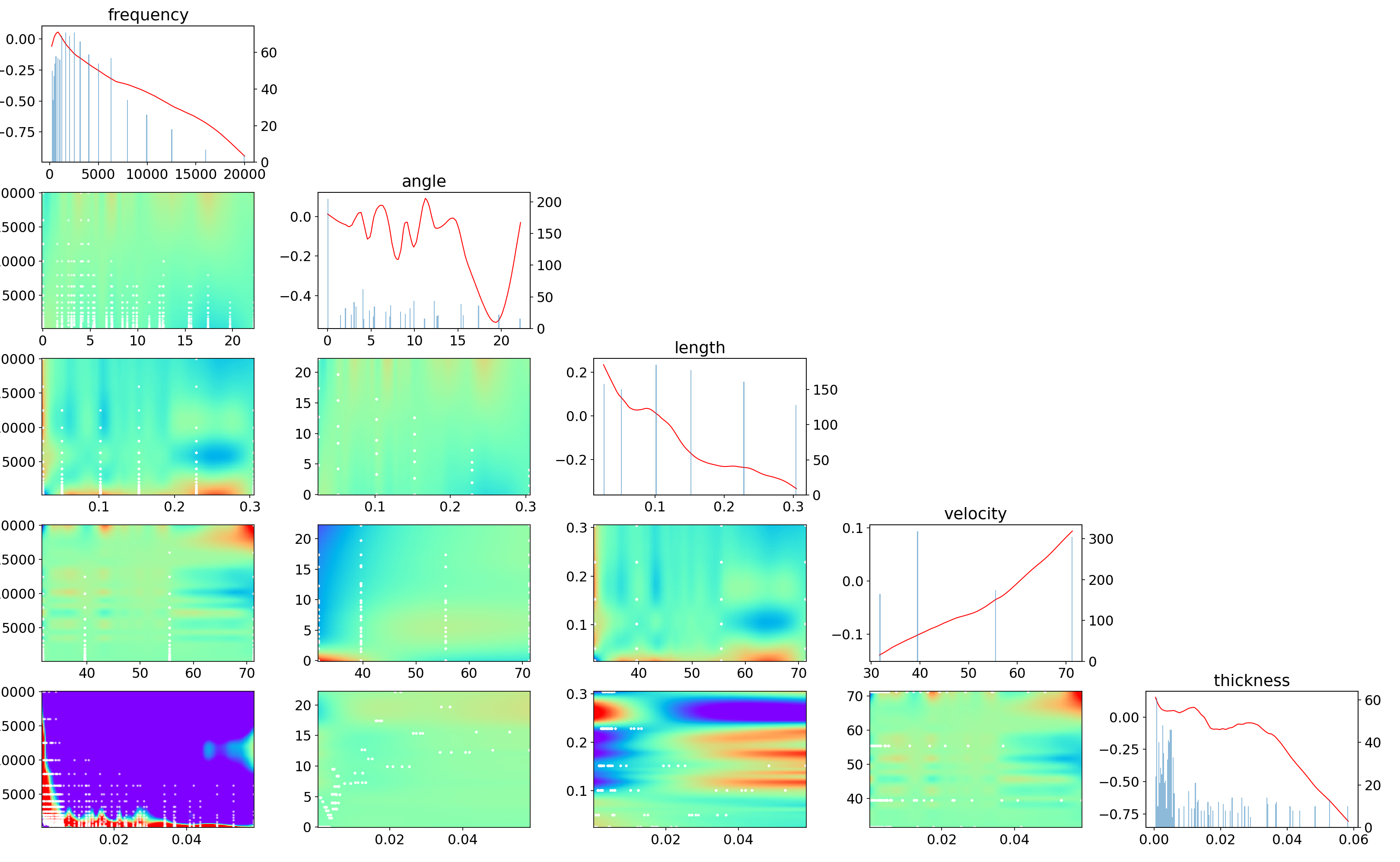}
    }\hfill
   \subfloat[fifa]{%
        \includegraphics[width=0.87\textwidth]{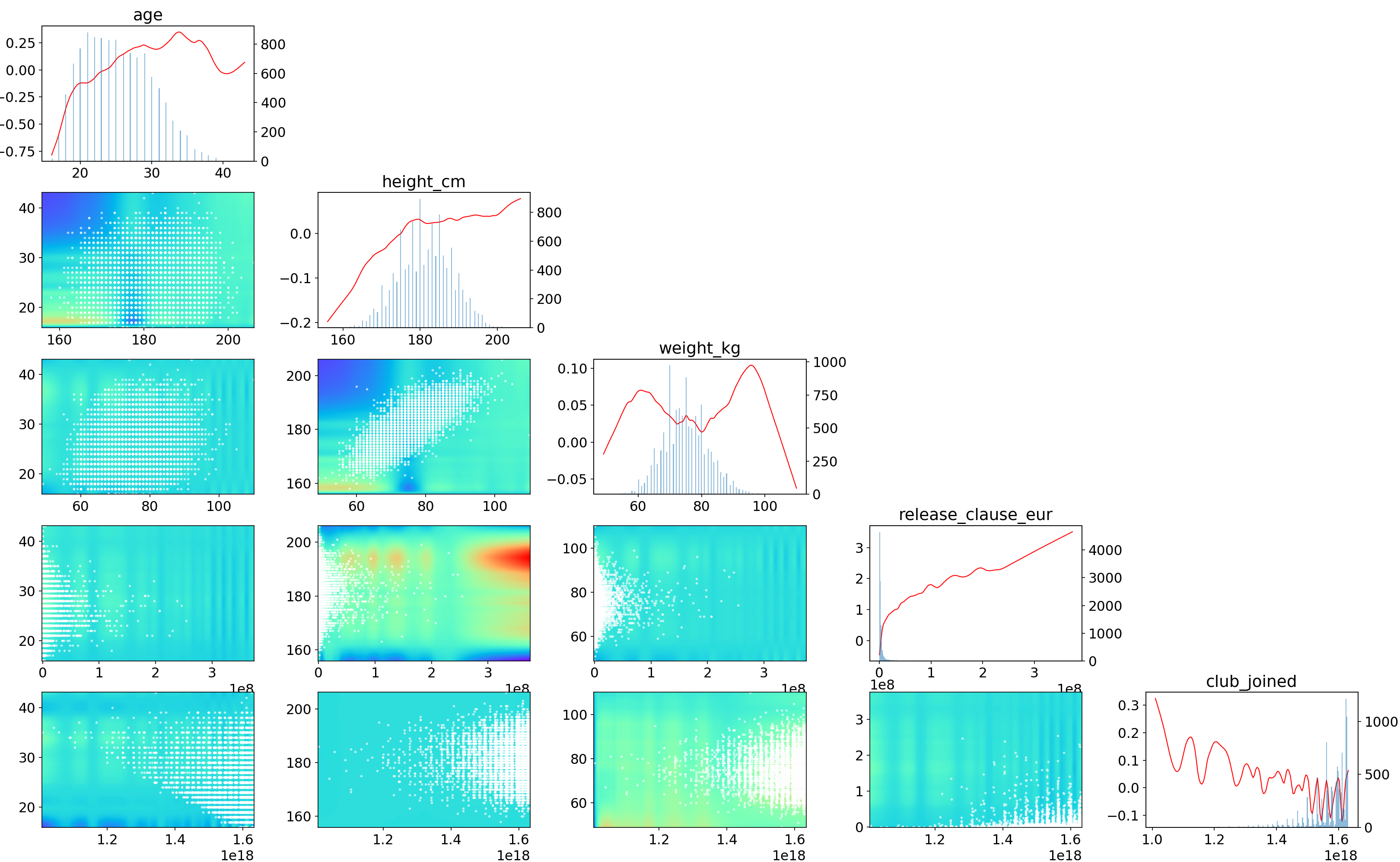}
    }\hfill
    \caption{Visualisation of the splines (1d and 2d) generated by the ${CoGA^2M}$ model on the a) airfoil\_self\_noise and b) fifa datasets. The 1d feature plots (labelled with the name of the corresponding feature) show the equivalent 1d spline (red curve). The left axis is the axis of the spline. The right is the axis of the histogram of the number of training data points over the range of the feature (blue histogram in the plot). This allows us to visualise directly the regions of the spline with and without good support. 
    The 2d heatmaps represent the arity 2 spline surfaces. The two inputs for each plot are the features of the 1d plots above and to the right of the 2d plot. So for example in panel A, the bottom left 2d plot would have as feature inputs the frequency (y axis) and thickness (x-axis) features. The white dots represent the coordinates (over the two features) where data points exist, again allowing us to directly visualise the support points of the surface.}
    \label{fig:arity2_splines}
\end{figure*}

%% file: Fig4.tex
\begin{figure}[h!]
    \centering
    \includegraphics[width=1\linewidth]{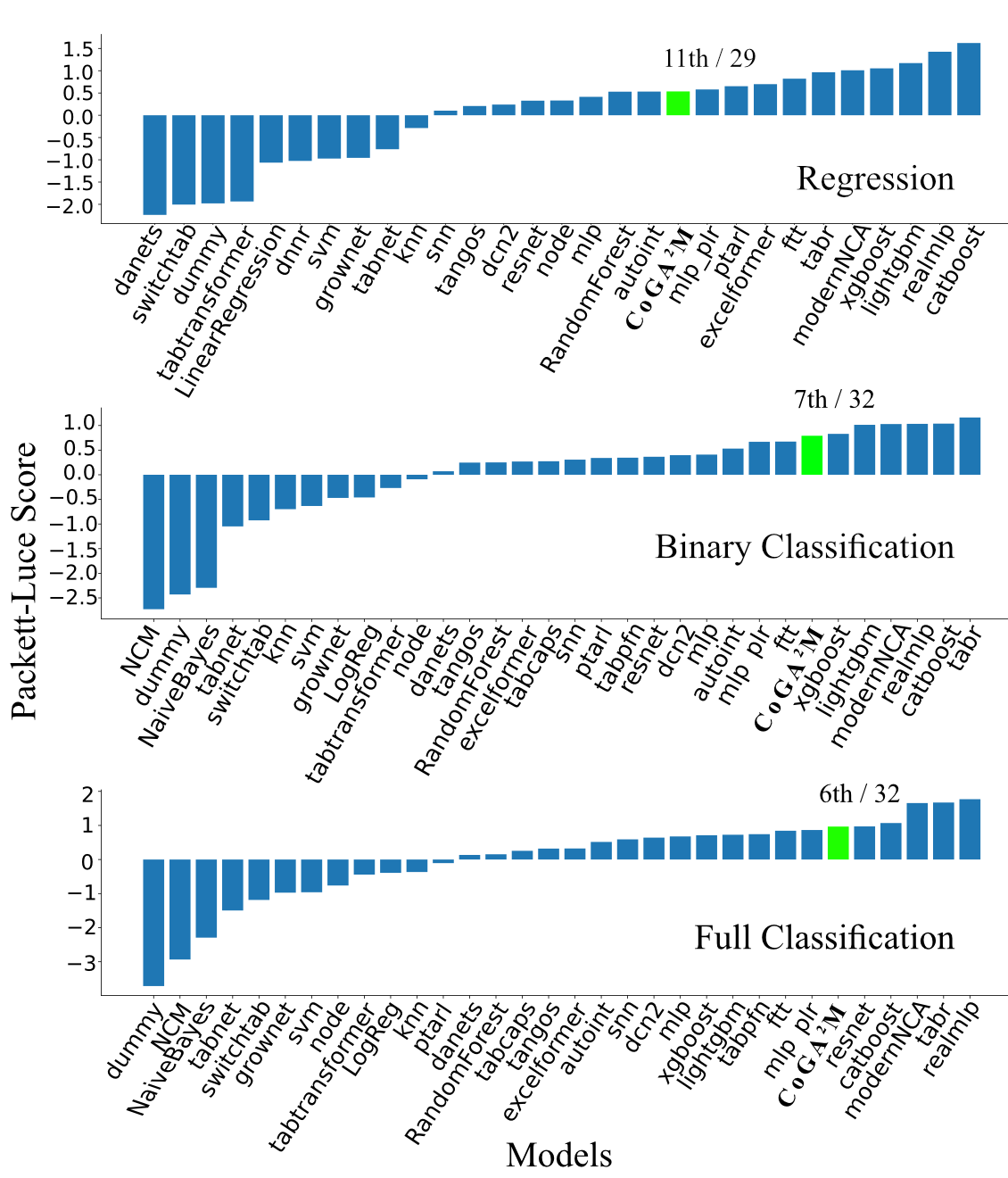}
    \caption{Results of the Packett-Luce score based ranking of the TALENT models plus ${CoGA^2M}$ for regression, binary classification and full classification datasets.}
    \label{fig:packett_luce_ranking}
\end{figure}
\label{talent_benchmark}

%% file: Fig5.tex
\begin{figure*}[h!]
    \centering
     \subfloat[Regression]{%
        \includegraphics[width=0.38\textwidth]{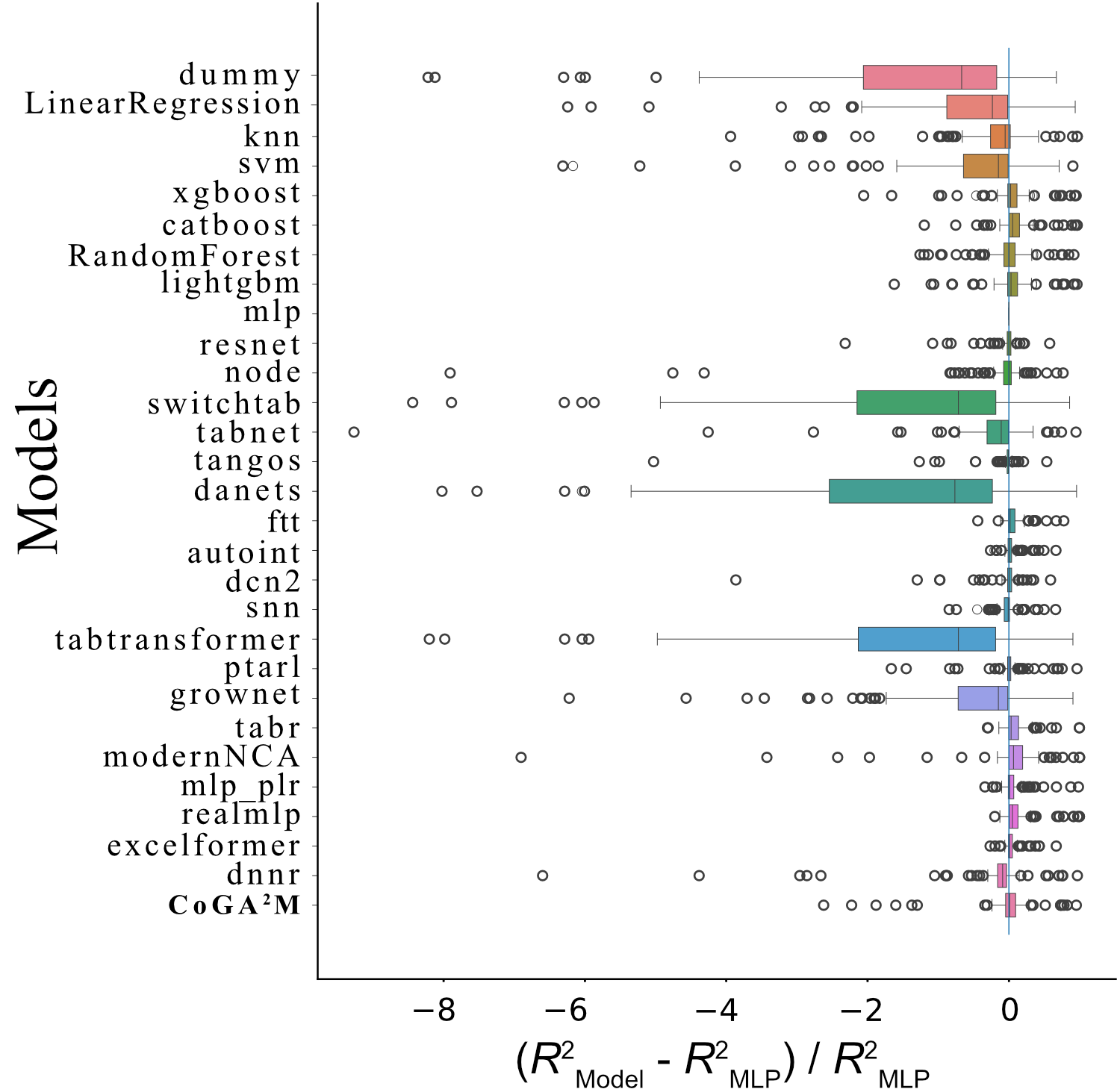}
    }
    \subfloat[Binary Classification]{%
        \includegraphics[width=0.35\textwidth]{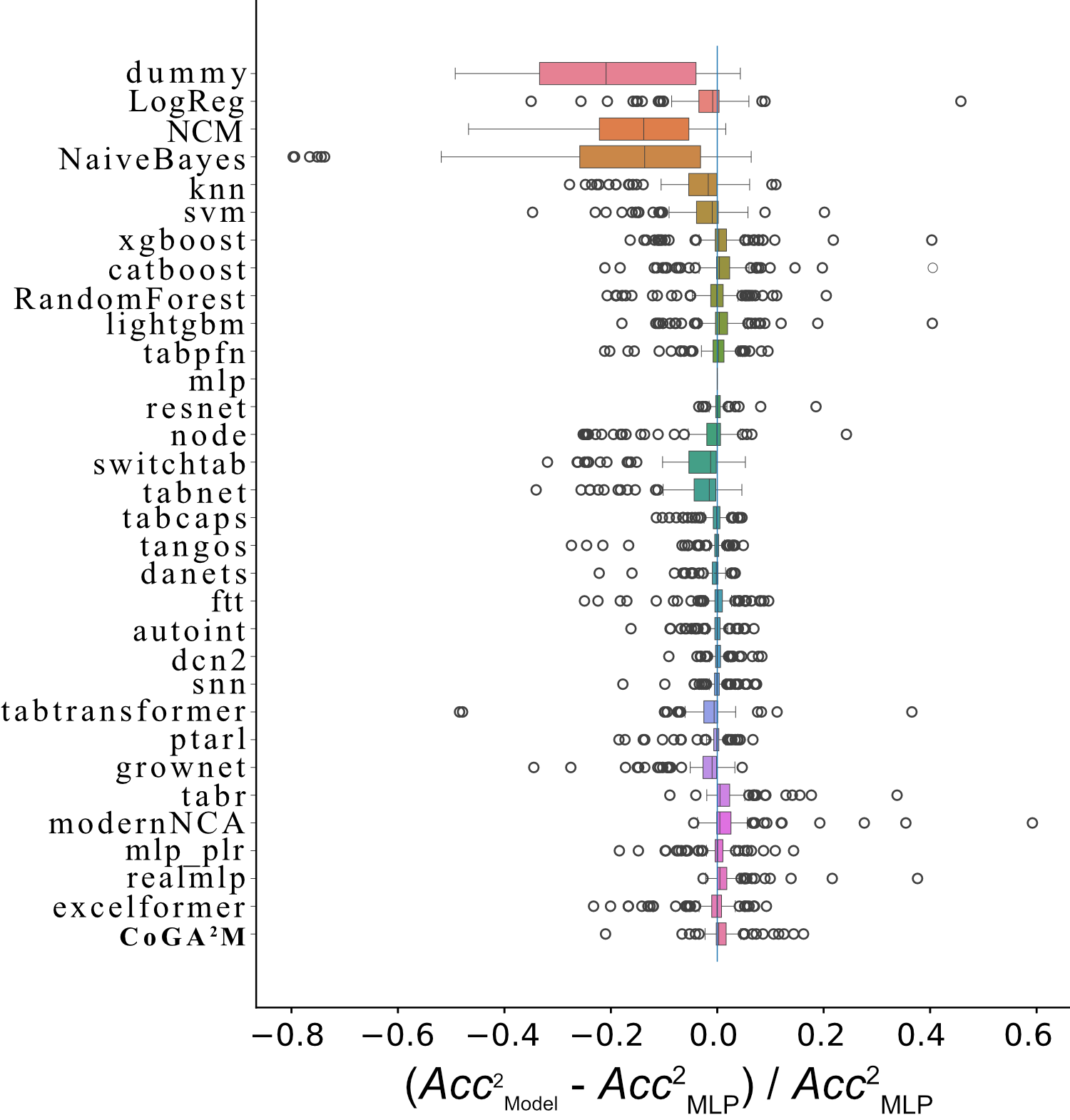}
    }
    \subfloat[Full Classification]{%
        \includegraphics[trim=9.3cm 0 0 0, clip, height=6.4cm]{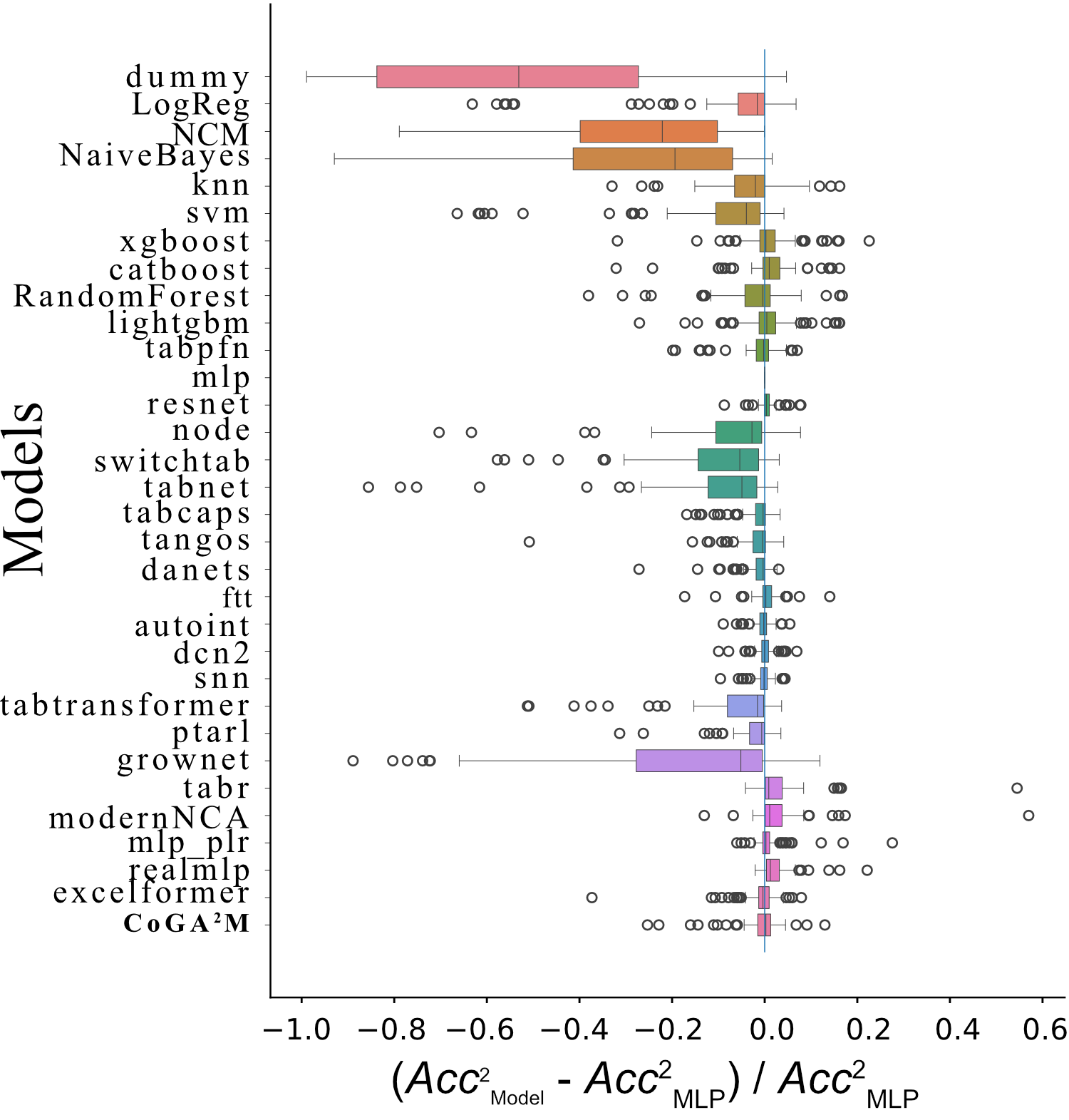}
    }\hfill
    \caption{Comparing all models' results with that of the MLP. For the regression models the metric used was the \(r^2\) while for the two classification types was Accuracy. The y-axis of models for the Full Classification is the same as the one for the Binary Classification.}
    \label{fig:mlp_ratios}
\end{figure*}

%% file: Fig6.tex
\begin{figure}[h!]
    \centering
    \includegraphics[width=1\linewidth]{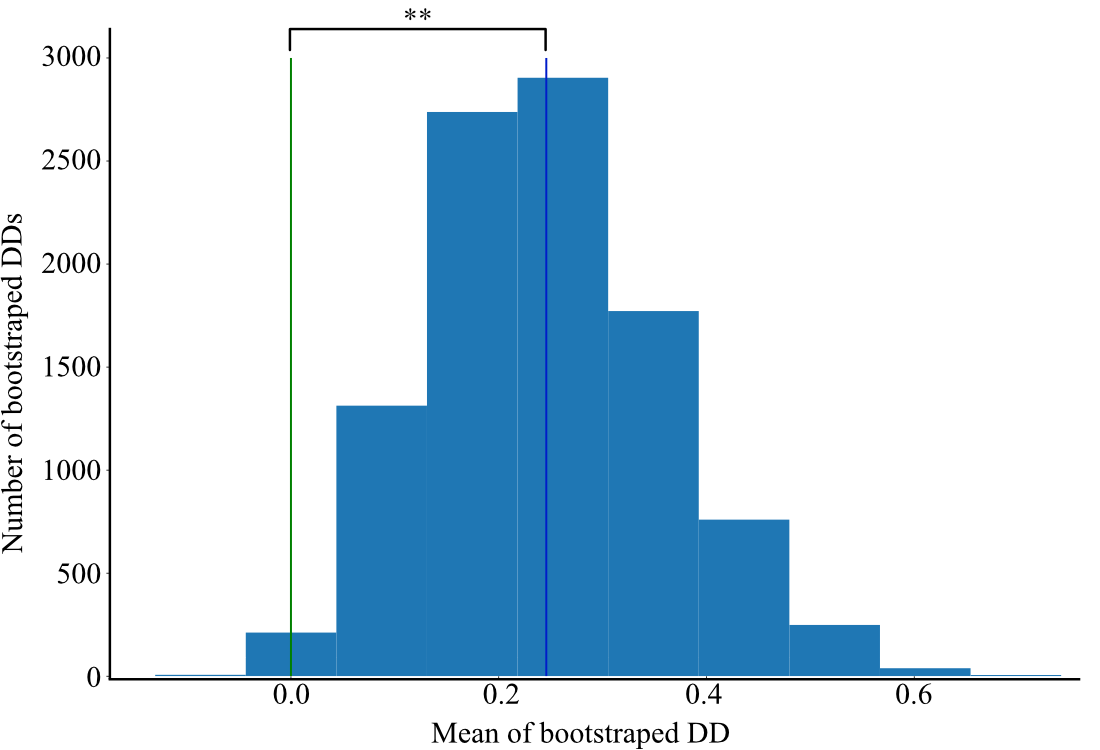}
    \caption{The histogram of the means of the bootstrap generated \(\Delta\Delta\) data showing that the distribution these data are drawn from is expected to have a larger than 0 mean.}
    \label{fig:DD_hist}
\end{figure}
\label{talent_benchmark}

%% file: SM_Fig1.tex
\begin{figure*}[h]
    \centering
    \subfloat[Angle]{%
        \includegraphics[trim=0 0 0 1cm, clip, width=0.95\textwidth]{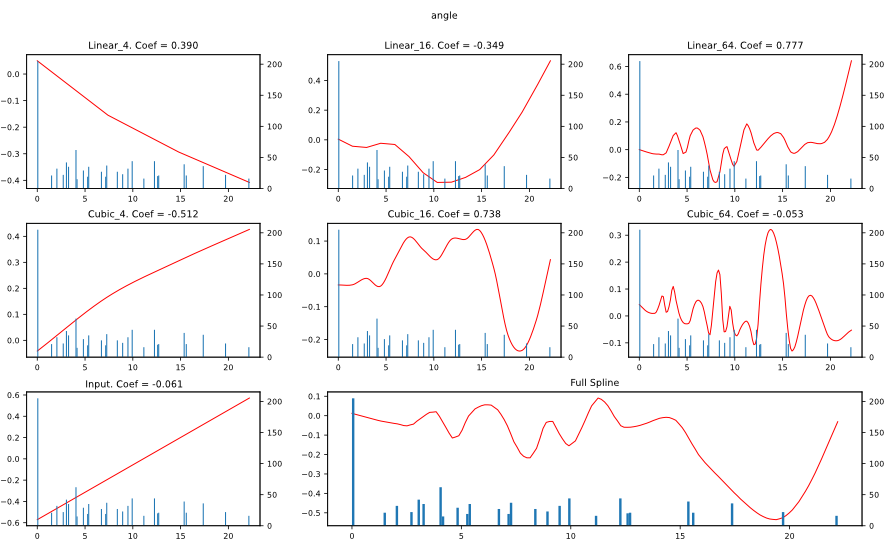}
    }\hfill
    \subfloat[Length]{%
        \includegraphics[trim=0 0 0 1cm, clip, width=0.95\textwidth]{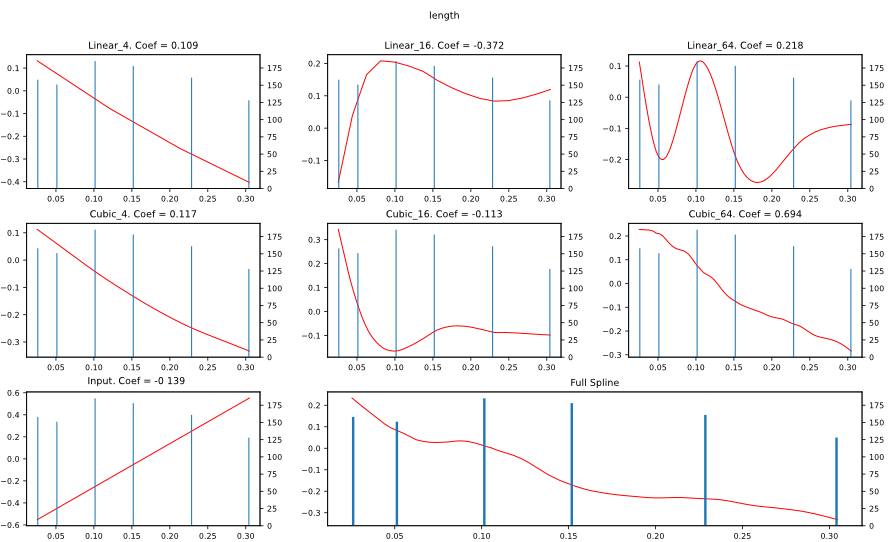}
    }
    \caption{Sub-splines for the 1d splines of the airfoil\_self\_noise dataset.}
    \label{fig:full_airfoil}
\end{figure*}

\clearpage

\begin{figure*}[h]
    \centering
    \ContinuedFloat
    \subfloat[Velocity]{%
        \includegraphics[trim=0 0 0 1cm, clip, width=0.95\textwidth]{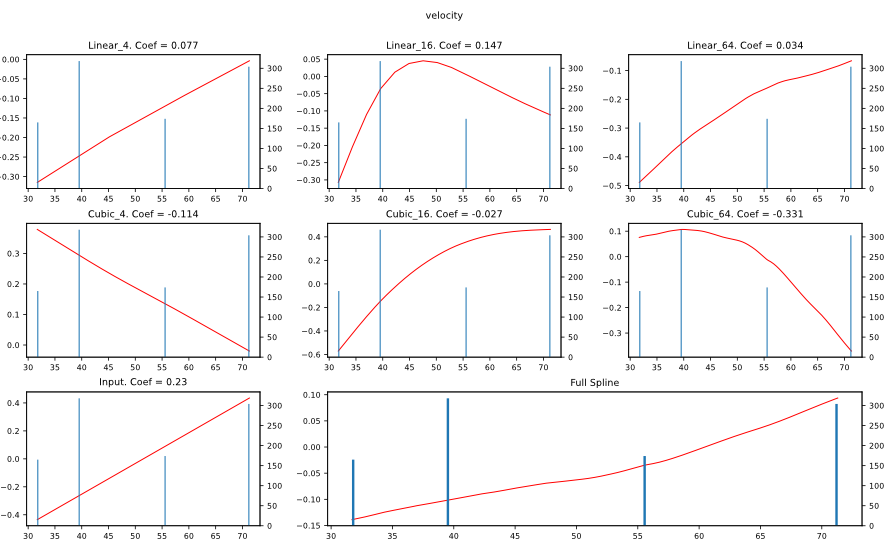}
    }\qquad
    \subfloat[Thickness]{%
        \includegraphics[trim=0 0 0 1cm, clip, width=0.95\textwidth]{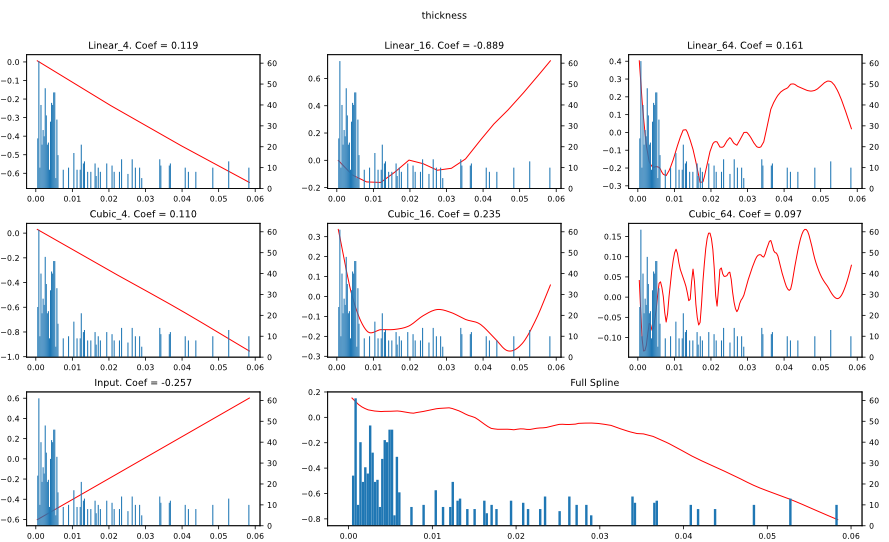}
    }\qquad
    \caption{Sub-splines for the 1d splines of the airfoil\_self\_noise dataset}
    \label{fig:full_airfoil}
\end{figure*}

\clearpage
\begin{figure*}[h]
    \centering
    \ContinuedFloat
    \subfloat[Frequency]{%
        \includegraphics[trim=0 0 0 1cm, clip, width=0.95\textwidth]{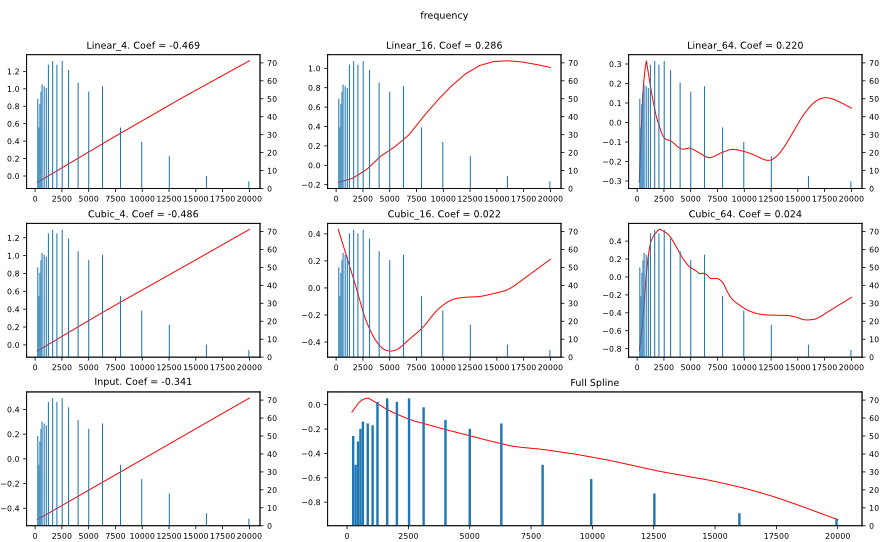}
    }
    \caption{Sub-splines for the 1d splines of the airfoil\_self\_noise dataset. For each feature there are 7 sub-splines linearly composed through learned weights (Coef. at the top of each sub-spline plot) to the final feature spline (bottom left of each group).The red lines are the sub-splines while the blue histogram (left y-axis) shows the data distribution).}
    \label{fig:full_airfoil}
\end{figure*}

%% file: SM_Fig2.tex
\clearpage
\begin{figure*}[h]
    \centering
    \subfloat[Height\_cm]{%
        \includegraphics[trim=0 0 0 1cm, clip, width=0.95\textwidth]{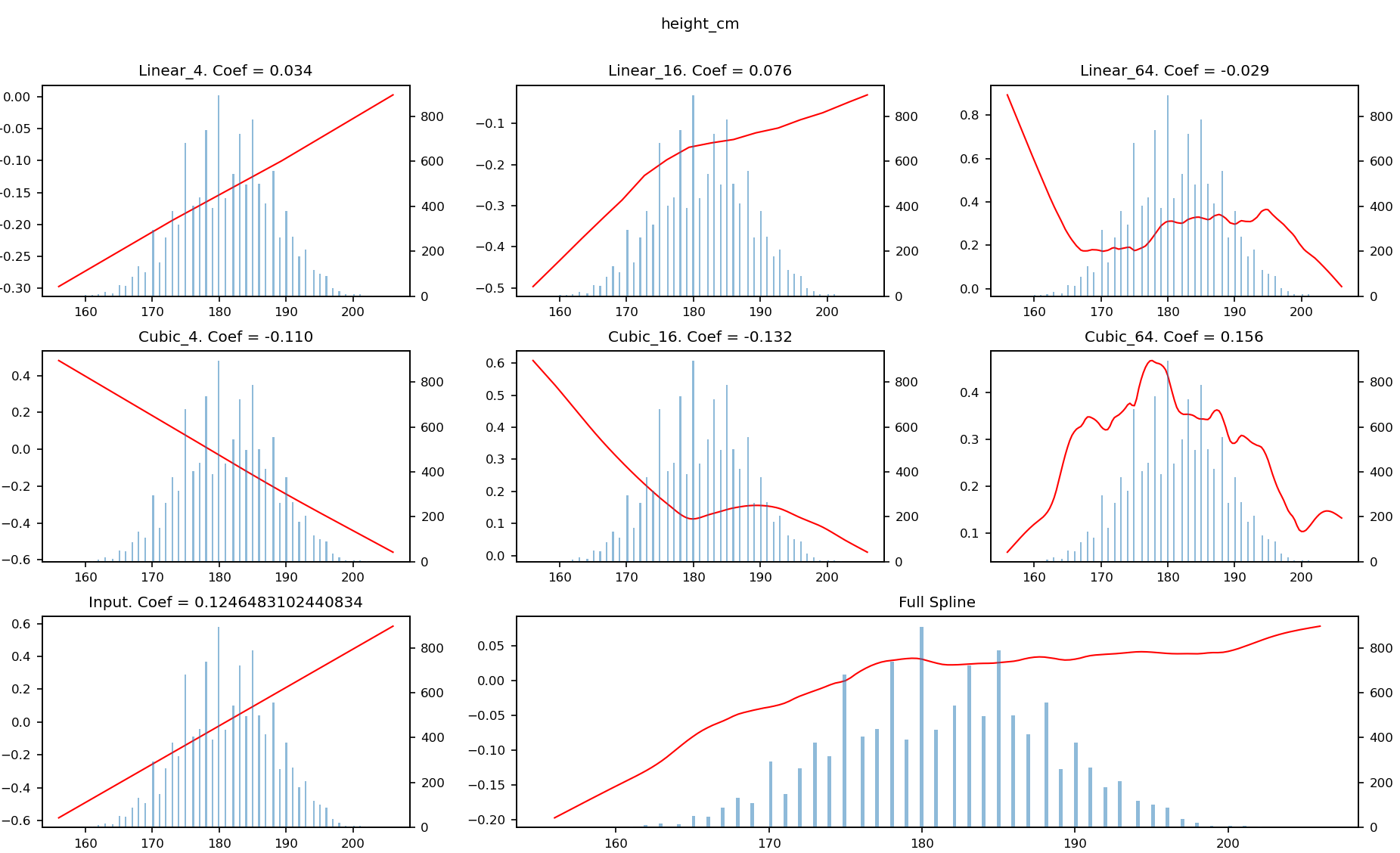}
    }\qquad
    \subfloat[Weight\_kg]{%
        \includegraphics[trim=0 0 0 1cm, clip, width=0.95\textwidth]{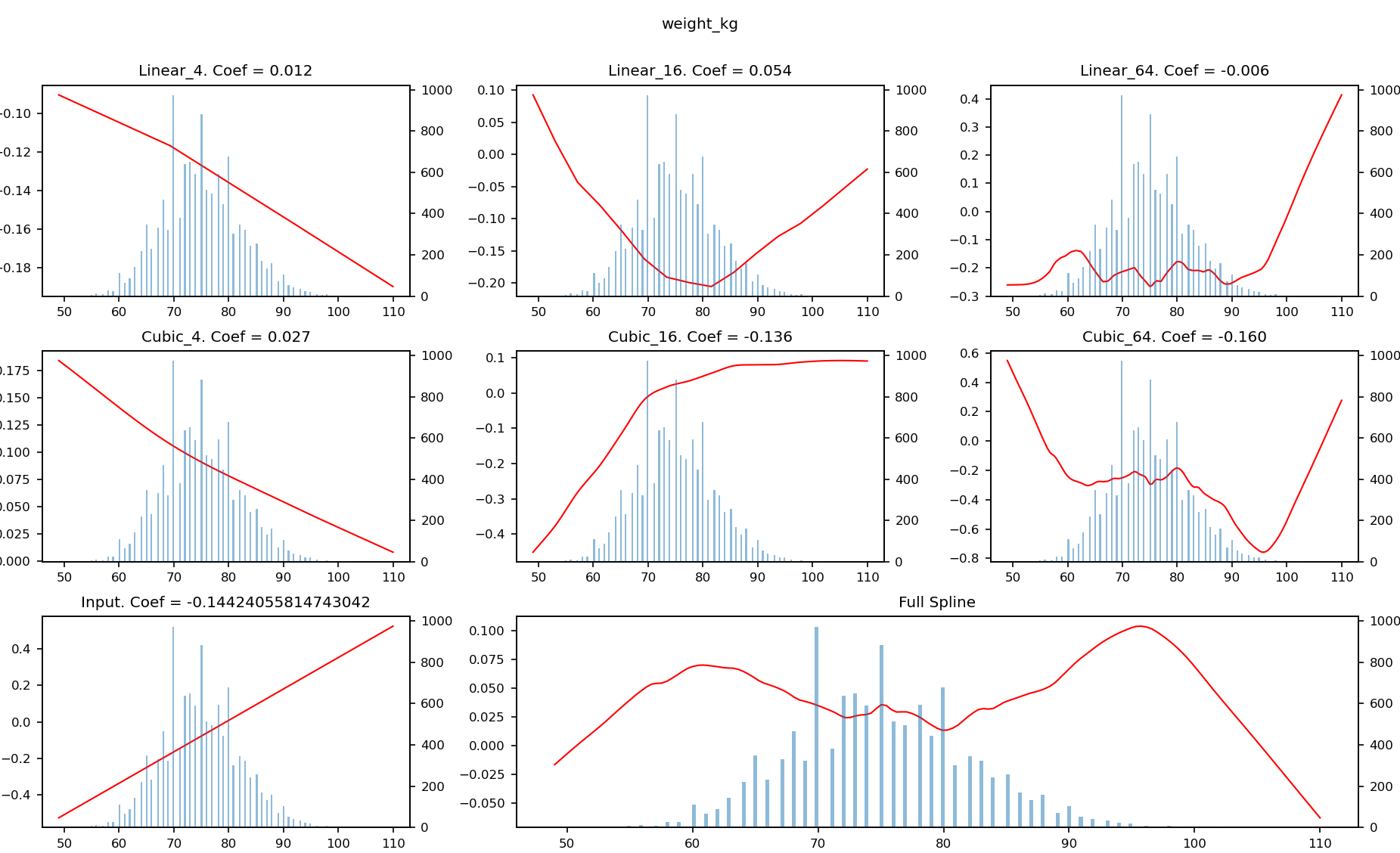}
    }\qquad
    \caption{Sub-splines for the 1d splines of the fifa dataset}
    \label{fig:full_fifa}
\end{figure*}

\clearpage
\begin{figure*}[h]
    \centering
    \ContinuedFloat
    \subfloat[Release\_clause\_eur]{%
        \includegraphics[trim=0 0 0 1cm, clip, width=0.95\textwidth]{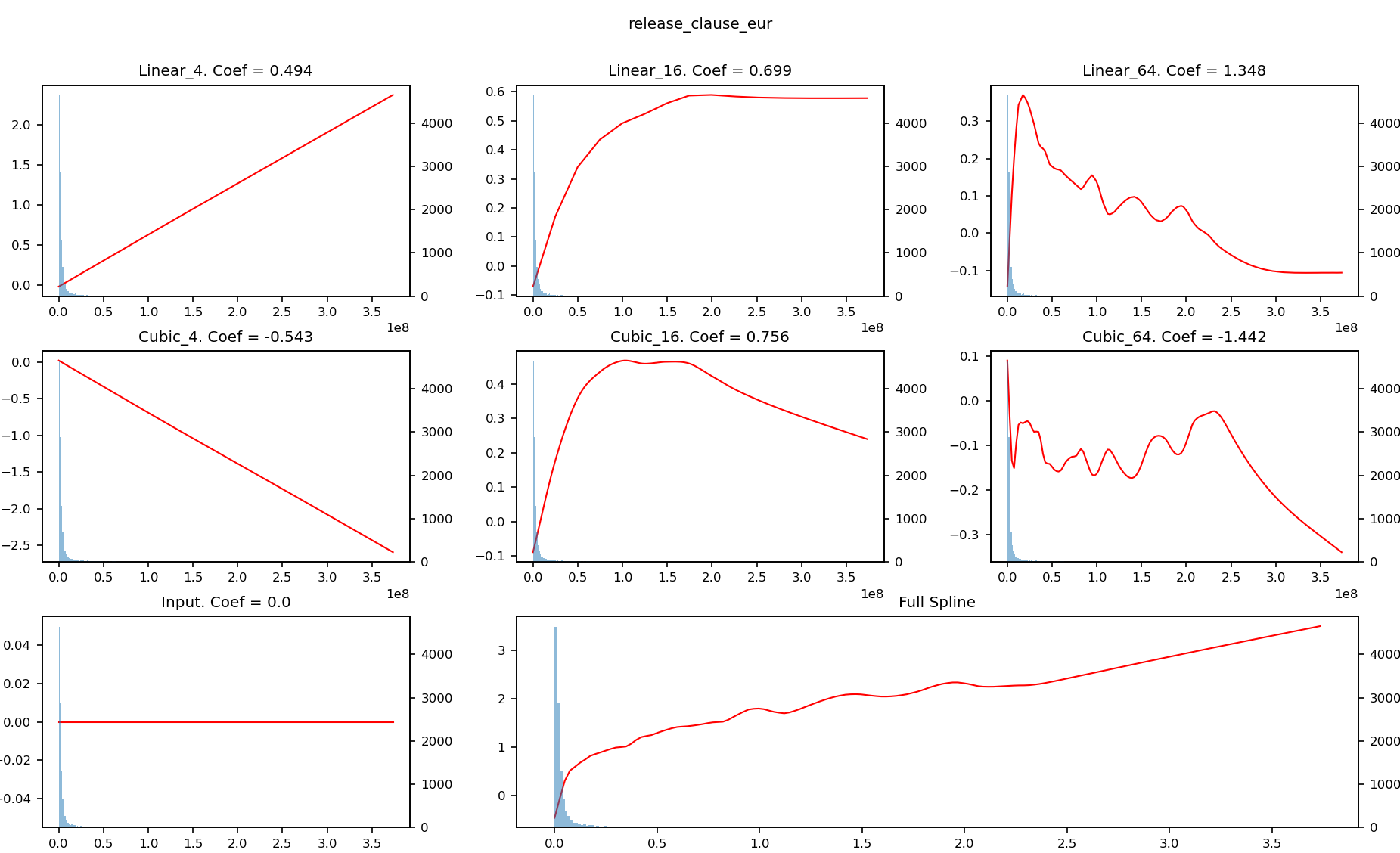}
    }\qquad
    \subfloat[Club\_joined]{%
        \includegraphics[trim=0 0 0 1cm, clip, width=0.95\textwidth]{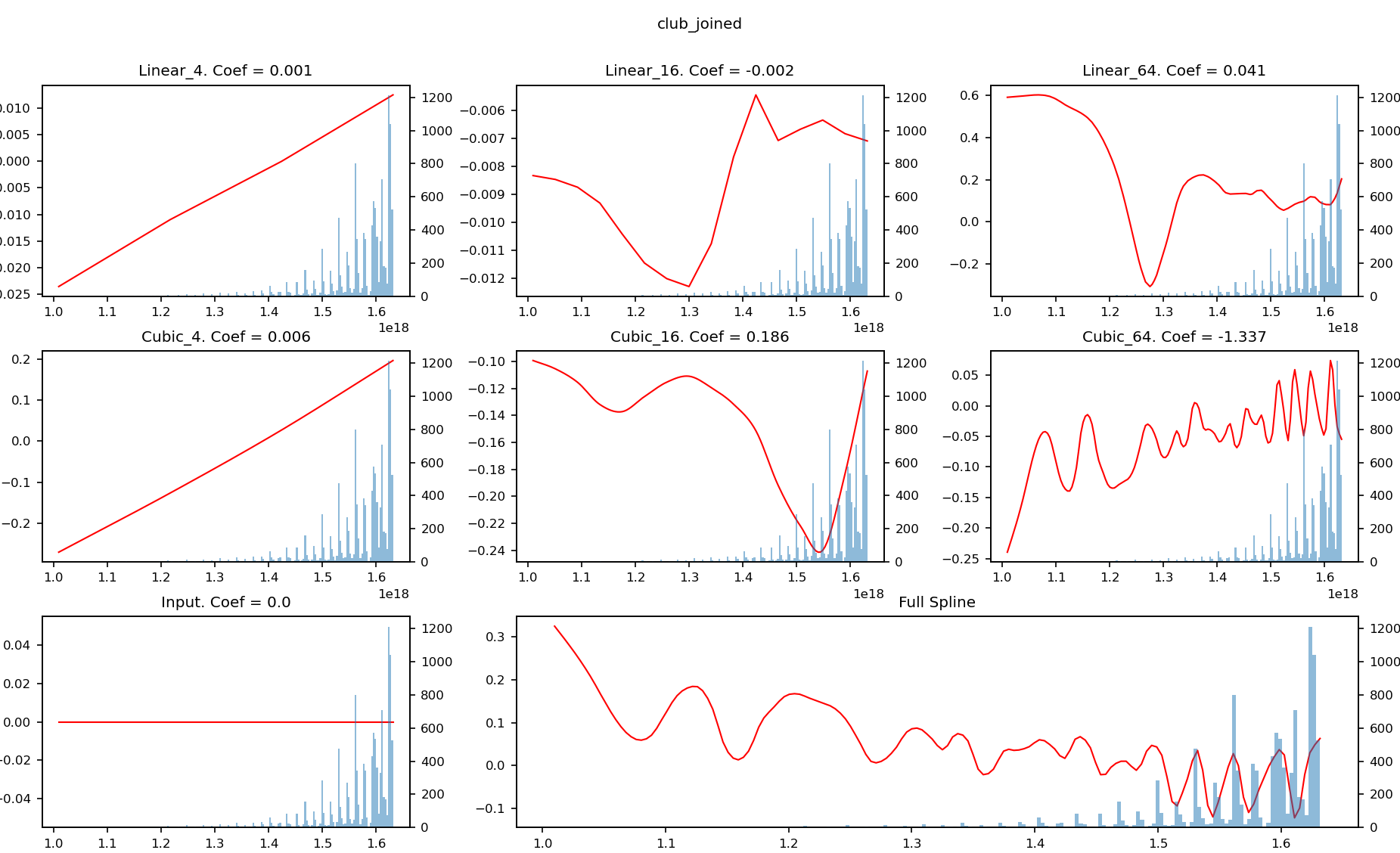}
    }
    \caption{Sub-splines for the 1d splines of the fifa dataset}
    \label{fig:full_fifa}
\end{figure*}

\clearpage

\begin{figure*}[h]
    \centering
    \ContinuedFloat
    \subfloat[age]{%
        \includegraphics[trim=0 0 0 1cm, clip, width=0.95\textwidth]{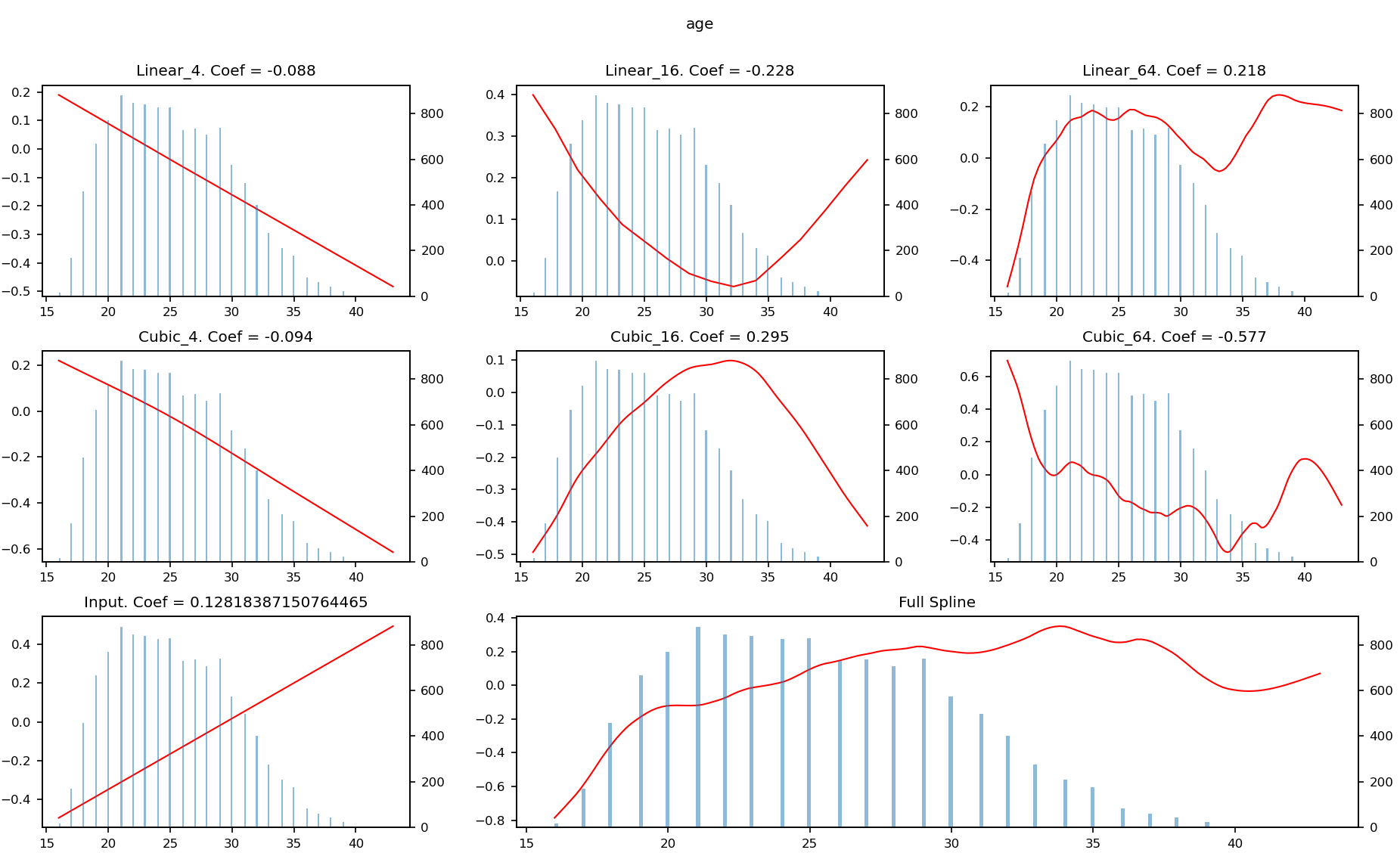}
    }\qquad
    \caption{Sub-splines for the 1d splines of the fifa dataset. For each feature there are 7 sub-splines linearly composed through learned weights (Coef. at the top of each sub-spline plot) to the final feature spline (bottom left of each group).The red lines are the sub-splines while the blue histogram (left y-axis) shows the data distribution)}
    \label{fig:full_fifa}
\end{figure*}

%% file: SM_fig3.tex
\begin{figure*}[h!t]
    \centering
     \subfloat[stock]{%
        \includegraphics[width=0.9\textwidth]{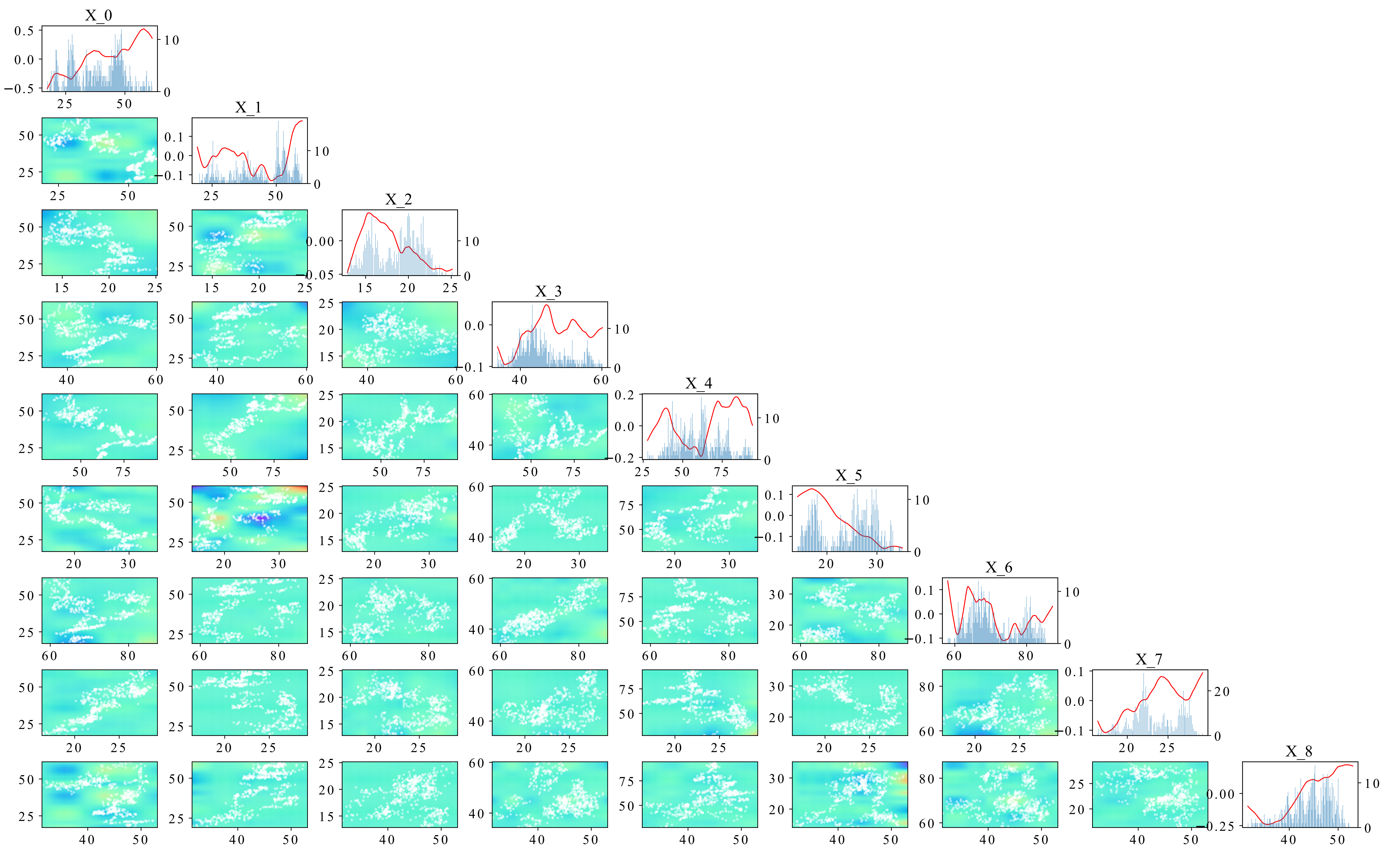}
    }\hfill
   \subfloat[concrete\_compressive\_strength]{%
        \includegraphics[width=0.9\textwidth]{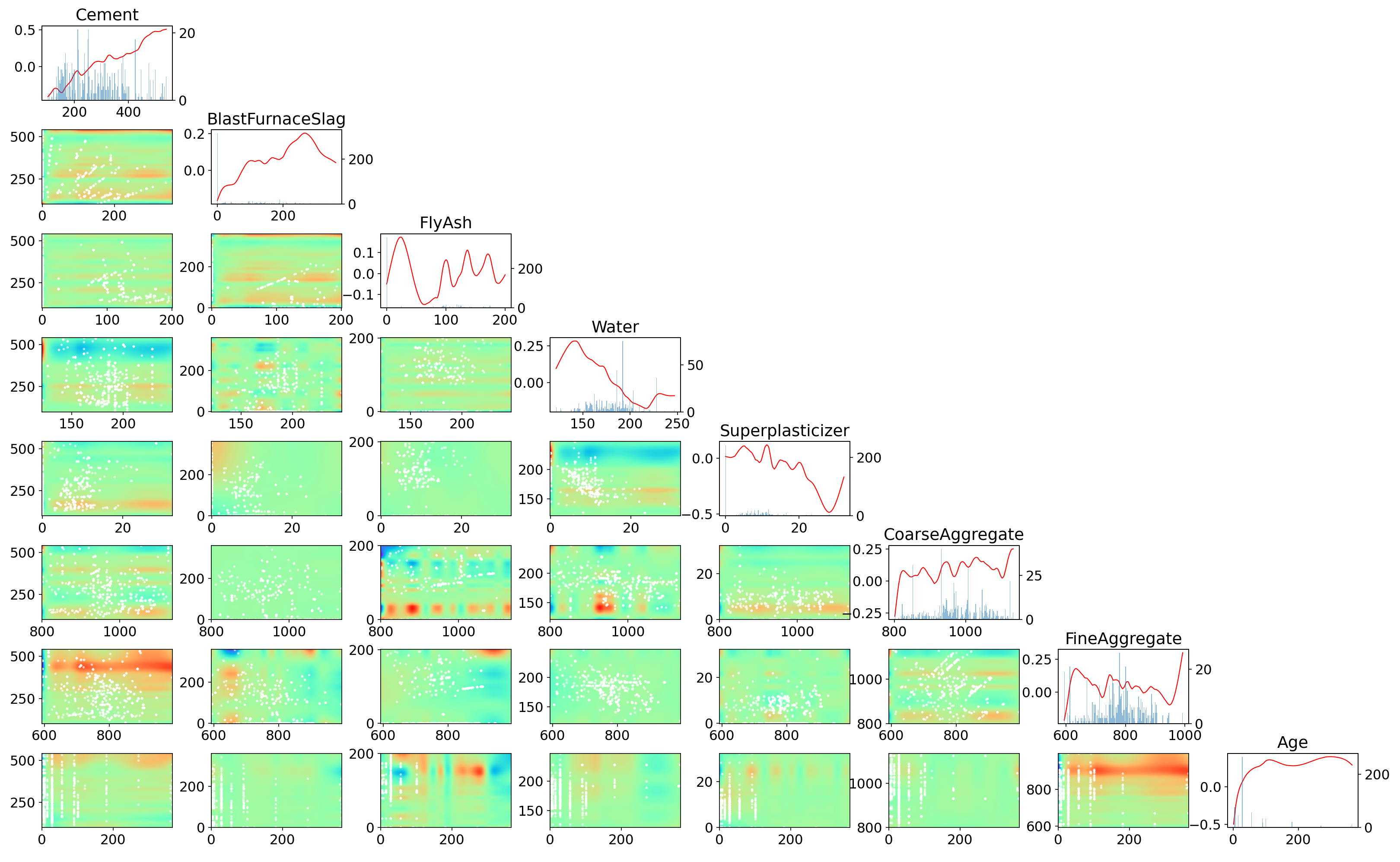}
    }\hfill
    \caption{Visualisation of the splines (1d and 2d) generated by the ${CoGA^2M}$ model on the (a) stock and (b) concrete\_compressive\_strength. See Fig. 2 for details.}
    \label{fig:arity2_splines_sm}
\end{figure*}